\documentclass[10pt,twocolumn,letterpaper]{article}

\usepackage{iccv}              

\usepackage{makecell}
\usepackage{multirow}
\usepackage{bm}
\usepackage{pifont}
\usepackage{algorithm}
\usepackage{algpseudocode}
\usepackage{balance}
\usepackage{subcaption}
\usepackage{graphicx}

\usepackage[textsize=tiny]{todonotes}
\DeclareMathOperator*{\argmax}{arg\,max}

\newcommand{\cmark}{\ding{51}}  
\newcommand{\xmark}{\ding{55}}  

\usepackage{graphicx}
\usepackage{color} 

\definecolor{Gray}{gray}{0.9}
\definecolor{Better}{rgb}{0.18, 0.407, 0.266}
\definecolor{Worse}{rgb}{0.35, 0.35, 0.35}
\definecolor{drakgreen}{rgb}{0.38, 0.67, 0.38}
\definecolor{drakpurple}{rgb}{0.38, 0.27, 0.61}
\definecolor{granate}{rgb}{0.64, 0.16, 0.16}

\definecolor{ours_color}{rgb}{0.92, 0.97, 1}

\newcommand{\Hquad}{\hspace{0.2em}} 
\newcommand{\mypar}[1]{\noindent\textbf{#1}\Hquad}

\definecolor{darkblue}{rgb}{0.21,0.49,0.74}

\usepackage{arydshln}
\usepackage{marvosym}

\newcommand{\ourcell}{\cellcolor{ours_color}}

\definecolor{iccvblue}{rgb}{0.21,0.49,0.74}

\usepackage[pagebackref,breaklinks,colorlinks,allcolors=iccvblue]{hyperref}
\usepackage{bbm}
\usepackage{dsfont}

\newcommand\blfootnote[1]{%
\begingroup
\renewcommand\thefootnote{}\footnote{#1}%
\addtocounter{footnote}{-1}%
\endgroup
}

\def\paperID{10053} 
\def\confName{ICCV}
\def\confYear{2025}

\title{\ours: Learning Vision-Language Uncertainties for Failure Prediction}

\author{
Marc Lafon$^{\star,1}$
\and
Yannis Karmim$^{\star,1}$
\and 
Julio Silva-Rodr\'iguez$^3$
\and
Paul Couairon${^2}$
\and 
Clément Rambour$^{1,2}$
\and 
Raphaël Fournier-Sniehotta$^{1,2}$
\and 
Ismail Ben Ayed$^3$
\and 
Jose Dolz$^3$
\and 
Nicolas Thome$^{2,4}$\\
$^1$\normalsize{Conservatoire National des Arts et Métiers, CEDRIC, F-75141 Paris, France}\\
$^2$\normalsize{Sorbonne Université, CNRS, ISIR, F-75005 Paris, France}\\
$^3$\normalsize{ETS Montreal, Canada} \\
$^4$\normalsize{Institut universitaire de France (IUF)}
}

\newcommand{\ours}{ViLU\xspace}

\begin{document}
\maketitle

\begin{abstract}

Reliable Uncertainty Quantification (UQ) and failure prediction remain open challenges for Vision-Language Models (VLMs). We introduce \textbf{\ours}, a new \textbf{Vi}sion-\textbf{L}anguage \textbf{U}ncertainty quantification framework that contextualizes uncertainty estimates by leveraging all task-relevant textual representations. \ours constructs an uncertainty-aware multi-modal representation by integrating the visual embedding, the predicted textual embedding, and an image-conditioned textual representation via cross-attention. Unlike traditional UQ methods based on loss prediction, \ours trains an uncertainty predictor as a binary classifier to distinguish correct from incorrect predictions using a weighted binary cross-entropy loss, making it loss-agnostic. ~In particular, our proposed approach is well-suited for post-hoc settings, where only vision and text embeddings are available without direct access to the model itself. Extensive experiments on diverse datasets show the significant gains of our method compared to state-of-the-art failure prediction methods. We apply our method to standard classification datasets, such as ImageNet-1k, as well as large-scale image-caption datasets like CC12M and LAION-400M. Ablation studies highlight the critical role of our architecture and training in achieving effective uncertainty quantification. Our code is publicly available and can be found here: \href{https://github.com/ykrmm/ViLU}{ViLU Repository}. 
\blfootnote{$^\star$ \text{Equal contribution} }
\blfootnote{Corresponding author: marc.lafon@lecnam.net}

\end{abstract}

\section{Introduction}
\label{sec:intro}


Vision Language Models (VLMs)~\cite{clip, siglip, molmo, idefics2, blip2} are
highly popular foundation models pre-trained on large-scale image-text datasets,
\eg, LAION~\cite{laion400m}. They possess the appealing capability of performing zero-shot
image classification
, meaning they can classify samples into classes
that were not seen during training. 


Uncertainty quantification (UQ) is a fundamental challenge in deep learning and involves estimating the confidence of a model's predictions. This paper addresses the problem of reliable UQ for VLMs to detect their potential failures in downstream tasks. Reliable UQ is crucial in safety-critical domains and offers significant opportunities for various applications involving VLMs, such as failure prediction~\cite{mcp}, out-of-distribution detection~\cite{GenOOD}, active learning~\cite{Learning_loss_AL}, and reinforcement learning~\cite{Galdropout}, among others.

\begin{figure*}[!ht]
   \centering
   \includegraphics[width=0.95\linewidth]{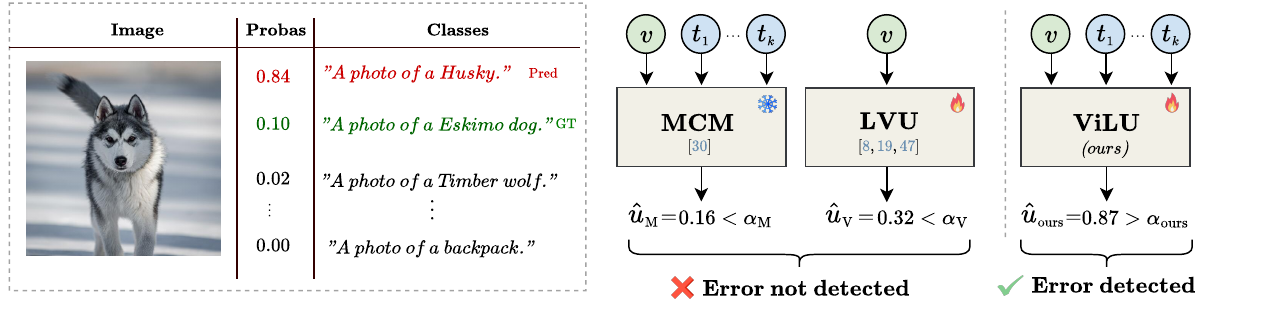}
    \vspace{-2em}
       \caption{\textbf{Motivation of \ours.} In zero-shot classification with VLMs, uncertainty may arise from both the image and target concept definitions. Given an ambiguity between several concepts, the vanilla Maximum Concept Maxing (MCM)~\cite{MCM}, can assign a high confidence to a wrong prediction, \eg an  "\texttt{American Eskimo dog}" wrongly classified as a "\texttt{Siberian husky}". Previous methods based on Learning Visual Uncertainties (LVU)~\cite{confidnet, Learning_loss_AL, pretrained_visual_uncertainties} do not account for ambiguity between concepts and often fail, whereas \ours captures a fine-grained uncertainty by contextualizing the image within the spectrum of possible textual concepts.}
   \label{fig:intro}
   \vspace{-0.3cm}
\end{figure*}

The vanilla method for UQ with VLMs is the Maximum Concept Matching (MCM) score~\cite{MCM}, a direct extension of Maximum Class Probability (MCP)~\cite{mcp} for classification. Although MCM is a strong baseline, it suffers from fundamental drawbacks: by design, it assigns high confidence to failures and struggles with fine-grained concepts. As illustrated in \cref{fig:intro}, the VLM misclassifies the ``\texttt{Eskimo dog}" image as a ``\texttt{Siberian husky}", and the high MCM score prevents the detection of the error.

In classification, learning-based methods have also been explored for failure prediction.
~In particular, a few deep learning models designed to predict the classifier's loss and dedicated to learning visual uncertainties (LVU) have been proposed~\cite{confidnet, Learning_loss_AL, pretrained_visual_uncertainties}. However, when applied to VLMs, these methods do not model the relationships between downstream concepts, intrinsically limiting their failure prediction performance. The example in \cref{fig:intro} still receives a high confidence score with LVU methods. Furthermore, although recent methods have proposed specific calibration techniques for VLMs~\cite{murugesan2024robust, wang2024open, yoonc, oh2025towards, probvlm}, fewer works have focused on UQ solutions for VLMs in general, with the exception of the recent~\cite{BayesVLM}.


This paper introduces \ours, a post-hoc framework for learning \textbf{Vi}sion-\textbf{L}anguage \textbf{U}ncertainties and detecting VLMs' failure.
~The core idea in \ours
is to define a confidence score depending both on the visual input but also on the set of concepts that defines the downstream task, \eg classification or caption matching. By finely modeling the interaction between the image and the target concepts, \ours assigns a low confidence score to the misclassified example -- see \cref{fig:intro}. 

We summarize our key contributions as follows:  
\begin{itemize}[topsep=0pt, noitemsep]  
    \item \ours introduces a novel multi-modal uncertainty representation that integrates visual embeddings, predicted textual embeddings, and image-conditioned textual representations via a cross-attention module. This formulation enables the model to capture fine-grained ambiguities between the input image and candidate concepts, significantly improving failure prediction.  

    \item  We propose a dedicated uncertainty predictor that operates on this enriched representation and is trained to discriminate between correct and incorrect predictions using weighted binary cross-entropy (BCE). Unlike conventional loss-prediction-based UQ methods, \ours is fully loss-agnostic, making it particularly well-suited for post-hoc uncertainty estimation in black-box VLMs.  
\end{itemize}


We conduct an extensive experimental validation of {\ours} on various downstream classification datasets, as well as image-caption datasets such as CC12M and LAION-400M. First, we highlight that state-of-the-art methods struggle to outperform the MCM baseline, illustrating the difficulty of the task. In contrast, \ours delivers significant and consistent improvements over multiple baselines, including recent VLM-specific methods~\cite{BayesVLM}. Thorough ablation studies further validate our architectural choices and training design for optimal performance. 

\section{Related Work}
\label{sec:related_work}

\mypar{Vision-language models (VLMs)} have gained popularity for aligning image and text representations \cite{clip, siglip, molmo, idefics2, blip2}, achieving unprecedented zero-shot performance by jointly learning a shared vision-text embedding space on large-scale web datasets \cite{laion400m, obelics, molmo}. They have 
been applied in a plethora of domains 
including image classification \cite{gallop, conditional_prompt_learning, prompt_vlm}, open-vocabulary segmentation \cite{fc_clip, maskclip, sclip, proxyclip} and cross-modal retrieval \cite{vop, clip_hitchhiker}. Nevertheless, 
VLMs align deterministic text and image representations without accounting for uncertainty in zero-shot predictions \cite{probvlm, BayesVLM}. While simple uncertainty quantification methods exist, \eg MCM \cite{MCM}, a robust approach for failure prediction remains lacking. We address this gap with an effective UQ method for failure prediction of VLMs.

\vspace{1mm}
\mypar{Uncertainty quantification (UQ) for failure prediction.} 
In classification, MCP \cite{mcp} is the vanilla method for UQ. However, by only considering the maximum over the predicted probabilities, MCP
~tends to overestimate confidence in failure cases~\cite{confidnet}. Another common approach is to use the Shannon entropy of the predicted softmax distribution~\cite{predictive_entropy}, but its invariance to label permutations \cite{predictive_entropy,doctor} limits its effectiveness for failure detection. Recently, Doctor~\cite{doctor} was proposed to refine the Shannon entropy~\cite{predictive_entropy} using Rényi entropy, while Rel-U~\cite{relu} further extended it by incorporating a learned distance matrix to model class relationships from classifier predictions. Such methods, however, suffer from a limited expressiveness in their UQ models, and struggle to capture finer-grained uncertainties. Another line of work is to learn the classifier's loss using deep neural networks~\cite{confidnet, Learning_loss_AL, pretrained_visual_uncertainties}. \cite{Learning_loss_AL} estimates model loss to improve active learning, while \cite{confidnet} learns the cross-entropy loss for classification and segmentation. More recently, \cite{pretrained_visual_uncertainties} applies this approach for large-scale UQ with vision transformers. However, these methods are limited to learning visual uncertainties (LVU), and generalize poorly to VLMs as they do not account for ambiguity in the language modality. To overcome this limitation, we propose an efficient UQ method for frozen, pre-trained VLMs and show that directly predicting failures outperforms loss prediction.


\vspace{1mm}
\mypar{VLMs' UQ.} MCP can be easily adapted for UQ of VLMs by leveraging their zero-shot probabilities, leading to the Maximum Concept Matching (MCM) method \cite{MCM}. However, MCM inherits MCP’s limitations for failure prediction, especially its overconfidence for errors. However, VLMs' overconfidence is generally less pronounced than for standard classifiers \cite{calibration_modern, calibration_closer_look}. As a result, when the VLM’s downstream accuracy is sufficiently high, MCM remains a strong baseline for failure prediction. Several works extend VLMs for UQ in cross-modal retrieval tasks~\cite{chun2021probabilistic, li2022a, Neculai2022ProbabilisticCE, improved_probabilistic}, often by learning probabilistic embeddings for each modality. However, most require retraining both visual and textual backbones, limiting their practicality
~\cite{improved_probabilistic, chun2021probabilistic}. To address this, ProbVLM~\cite{probvlm} learns distribution parameters over embeddings via adapters, but overlooks cross-modal similarity scores, which limits its effectiveness for failure prediction. BayesVLM~\cite{BayesVLM} applies a Laplace approximation to model uncertainty over similarities post-hoc. While both capture vision-language uncertainty, their objectives are not tailored for failure prediction, which limits their performance compared to our approach.

\begin{figure*}[!t]
   \centering
   \includegraphics[width=0.95\linewidth]{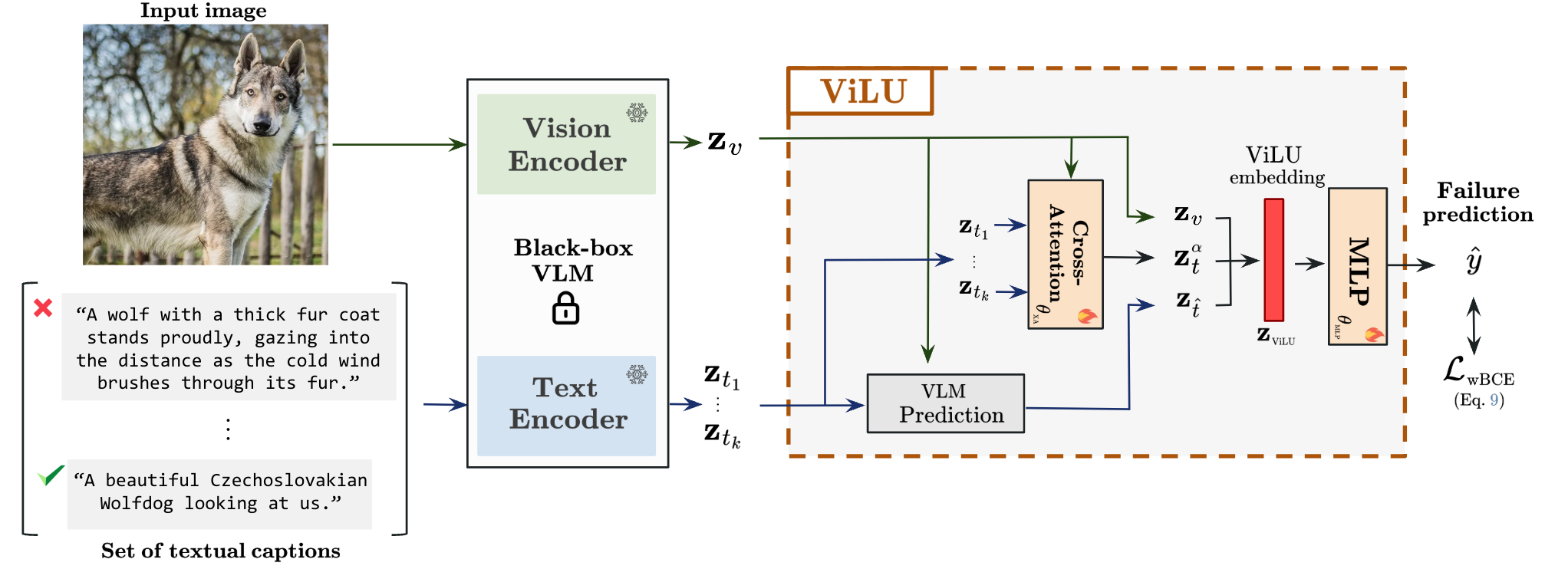}
   \caption{\textbf{Overview of \ours:} A learning-based \textbf{Vi}sion-\textbf{L}anguage \textbf{U}ncertainty quantification framework for VLM failure prediction. The key strength of \ours lies in its ability to contextualize uncertainty estimates by leveraging all textual representations relevant to the task (see \cref{sec:methodo}). It constructs an uncertainty-aware representation by combining the visual embedding $\bm{z}_{v}$, the predicted textual embedding $\bm{z}_{\hat{t}}$, and an image-conditioned textual representation $\bm{z}^{\alpha}_{t}$ obtained via cross-attention (see \cref{sec:archi}). Instead of relying on loss prediction, \ours trains an uncertainty predictor as a binary classifier to distinguish between correct and incorrect predictions (see \cref{sec:training}). \ours is a post-hoc approach that can be efficiently deployed on top of any pre-trained VLM without accessing its weights (black box) and supports both image-caption and image-label tasks.}
   \label{fig:method}
   \vspace{-0.6em}
\end{figure*}

\section{Background}
\label{sec:background}

\subsection{Contrastive vision-language models}
\label{background_zeroshot}

\noindent\textbf{Zero-shot predictions.} Consider a particular dataset $\mathcal{D} = \{(\bm{x}_i, t_i)\}_{i=1}^N$, where each image $\bm{x}_i \in \mathcal{X}$ is paired with a class name or caption $\bm{t}_i \in \mathcal{T}$.
~Contrastive VLMs are composed of a pre-trained vision encoder $f_{\mathcal{V}}(\cdot)$ that maps an input image $\bm{x}_i$ to a visual embedding $\bm{z}_{v_i} = f_{\mathcal{V}}(\bm{x}_i) \in \mathbb{R}^{d}$, and a text encoder $f_{\mathcal{T}}(\cdot)$ that embeds a textual input $\bm{t}_j$ into a representation $\bm{z}_{t_j} = f_{\mathcal{T}}(\bm{t}_j) \in \mathbb{R}^{d}$. This multi-modal joint embedding space is learned during pre-training by aligning paired concepts. Thus, it enables zero-shot image classification, by computing the probability $p(\bm{t}_j|\bm{x}_i)$ of image $\bm{x}_i$ being associated to caption ${\bm{t}_j}$ using the softmax of the similarities between the visual representations $\bm{z}_{v_i}$ and the embeddings $\bm{z}_{t_j}$ of a set of $K$ candidate textual concepts:
\vspace{-0.2em}
\begin{equation}
p(\bm{t}_j|\bm{x}_i) = \frac{\exp{({\bm{z}}_{v_i}^\top {\bm{z}}_{t_j} / \tau)}}{\sum_k \exp{({\bm{z}}_{v_i}^\top {\bm{z}}_{t_k} / \tau)}}
\end{equation}

\noindent where $\tau$ is a temperature parameter optimized during pre-training, and ${\bm{z}}_{v_i}$ and ${\bm{z}}_{t_j}$ are $\ell_{2}$-normalized embeddings.

\subsection{VLMs' failure prediction with MCM}

We aim to determine whether VLMs can recognize when their predictions are unreliable. This is captured by an uncertainty scoring function,  $u(\bm{x})$, where higher values indicate a greater likelihood of misclassification. 

\vspace{0.5em}
\mypar{Maximum Concept Matching (MCM) \cite{MCM}} 
 corresponds to the probability of the predicted caption for image $\bm{x}_i$, \ie, the one with the highest probability:
\begin{equation}
u_{\text{MCM}}(\bm{x}_i, \bm{t}_1, ..., \bm{t}_k) = 1 - \max_{j} p(\bm{t}_j|\bm{x}_i).
\label{eq:mcm}
\end{equation}

\noindent\textbf{Limitations.}  
Despite being a reasonable UQ method in coarse-grained classification tasks, 
$u_{\text{MCM}}$ has fundamental limitations. Firstly, the max operation in \cref{eq:mcm} by design assigns an overestimated UQ for incorrect predictions. Therefore, MCM performances drop when the zero-accuracy of the classifier is low, as shown in the experiments.
~Also, in fine-grained classification or open-vocabulary settings, visual and textual alignments become more dispersed, reducing MCM's reliability. This limitation is particularly problematic for modern VLMs, which are trained on large-scale datasets for open-vocabulary tasks.



\section{\ours model}
\label{sec:method}
This section presents our \ours framework for multi-modal failure prediction on VLMs, as illustrated in \cref{fig:method}. We first introduce our general post-hoc methodology
~in \cref{sec:methodo}, which enables  UQ on VLMs without modifying their internal parameters. We then describe our architecture in \cref{sec:archi}, detailing the design of \ours's embedding using image-text cross-attention, and of our failure classification head.
~Finally, we outline \ours's training procedure  in \cref{sec:training}, including our weighted cross-entropy loss that directly aligns with the failure prediction task.

\subsection{Methodology for UQ on VLMs}
\label{sec:methodo}
\mypar{Post-hoc Setting.}  We aim to design a discriminative and reliable UQ measure for pre-trained VLMs. To achieve this, we adopt a post-hoc approach, relying solely on visual and textual representations, \ie, $\mathcal{D} = \{(\bm{z}_{v_i}, \bm{z}_{t_i})\}_{i=1}^N$. This allows the UQ model to be easily integrated on top of a pre-trained VLM, providing uncertainty estimates without modifying internal representations, requiring fine-tuning, or depending on the loss function used during training.

\vspace{1mm}
\mypar{Learning vision-language uncertainties.} We propose leveraging the interactions between the \textit{visual modality} and the \textit{set of candidate concepts} to estimate uncertainty for failure prediction. Uncertainty in model predictions can arise from \textit{visual patterns} (\eg, low image quality, ambiguous features) or \textit{textual patterns}, which define concept distinctions. Additionally, it is shaped by cross-modal interactions. To model these interactions, we learn a global uncertainty representation $u_{\theta}(\cdot)$ that captures visual-textual interactions and their inherent uncertainty.  We adopt a \textit{data-driven approach}, training the uncertainty module to predict VLMs' misclassifications. Specifically, we frame our objective as a \textit{binary classification task}, where $u_{\theta}$ predicts whether an input will be misclassified by the VLM (see \cref{sec:training}). Formally, let us define our uncertainty module as:
\begin{align}
\begin{split}
    u_{\theta}:\quad \quad \mathbb{R}^{d}\times{\mathbb{R}^{K \times d}} &\quad \rightarrow \quad \quad \quad [0, 1]\\
                 (\bm{z}_{v}, \bm{z}_{t_1}, ..., \bm{z}_{t_K}) &\quad \rightarrow \quad u_{\theta}(\bm{z}_{v}, \bm{z}_{t_1}, ..., \bm{z}_{t_K})
\end{split}
\label{eq:vilu_score}
\end{align}

Note that our uncertainty function in \cref{eq:vilu_score} can handle varying values of $K$, as the number of candidate concepts may vary during inference.


\subsection{\ours's architecture}
\label{sec:archi}
This section details the architectural components of our uncertainty module $u_{\theta}$ and how a visual feature, $\bm{z}_{v}$, for a given sample and textual embeddings of $K$ target concepts, ${Z}_{t}={\{\bm{z}_{t_j}\}}_{1 \leq j \leq K}$, are combined to provide an expressive representation for failure detection.

\vspace{1mm}
\mypar{Vision-textual cross-attention.} To enable flexible representations of the visual and textual modalities, we employ a \emph{cross-attention module} $h_{\theta_{\text{XA}}}$ that produces an image-specific textual representation $\bm{z}^{\alpha}_{t}$ allowing to efficiently capture inherent cross-modal uncertainty:
\begin{equation}
\textstyle \bm{z}^{\alpha}_{t} = h_{\theta_{\text{XA}}}(\bm{z}_{v}, \bm{z}_{t_1}, ..., \bm{z}_{t_K})
\end{equation}

More specifically, the \emph{query} is the visual representation $\bm{z}_{v}$, and \emph{keys and values} are the $K$ textual embeddings ${Z}_{t}$. The cross-attention's output is obtained as:
\begin{align}
\begin{split}
\bm{\alpha} = \text{softmax}&\left((W_Q\bm{z}_{v})^\top ~ (W_K{Z}_{t}) / \sqrt{d} \right)\\
\bm{z}^{\alpha}_{t} &= \sum_{j=1}^{K} \bm{\alpha}_{j} (W_V\bm{z}_{t_j})\\
\end{split}
\end{align}
where $\bm{\alpha}$ are the attention weights, and  $W_Q$, $W_K$, and $W_V$ denote the projection matrices for the queries, keys, and values, respectively. This weighted textual embedding refines the textual context based on the model’s predicted distribution over the candidate captions, allowing for a more accurate uncertainty assessment. This contextualized representation can be computed for any number of concepts, ensuring a generic uncertainty module that remains well-defined across different concept sets.

\vspace{0.25em}
\mypar{ViLU embeddings.} To resolve vision-language ambiguity for failure prediction, our model must capture key information to distinguish correct from incorrect predictions. At a minimum, this requires the visual representation $\bm{z}_v$ and the textual embedding of the predicted caption $\bm{z}_{\hat{t}}$, allowing the model to approximate MCM's uncertainty estimator. However, this limited representation overlooks ambiguous alternatives, making error detection unreliable. To address this, we construct a rich vision-language uncertainty embedding $\bm{z}_{\text{\tiny{ViLU}}}=(\bm{z}_v, \bm{z}_{\hat{t}}, \bm{z}^{\alpha}_{t})$ by concatenating $\bm{z}_v$, $\bm{z}_{\hat{t}}$, and the cross-attention output $\bm{z}^{\alpha}_{t}$.   

\vspace{0.25em}
\mypar{Learning complex patterns for failure prediction.} Detecting misclassifications among numerous fine-grained concepts is challenging, as it requires capturing complex cross-modal relationships. To overcome this, we apply a non-linear transformation on our ViLU embeddings via a multi-layer perceptron (MLP), $g_{\theta_{\text{MLP}}}$, which enhances feature expressiveness and produces a scalar uncertainty estimate. Formally, uncertainty quantification is expressed as the predicted failure score $\hat{y}_i$:
\begin{equation}
\hat{y}_i = \sigma\bigl( g_{\theta_{\text{MLP}}}(\bm{z}_{\text{\tiny{ViLU}}}) \bigr),
\label{eq:vilu_failure-pred}
\end{equation}
where $\sigma$ denotes the sigmoid function, ensuring that $\hat{y}_i \rightarrow 1$ when the predicted zero-shot label is likely incorrect. Importantly, $g_{\theta_{\text{MLP}}}$ in \cref{eq:vilu_failure-pred} boils down to the unnormalized MCM score when using a bilinear form on $\bm{z}_{\text{\tiny{ViLU}}}$ (see \cref{sec:supp-vilu_details}). Our failure predictor is thus a \textit{consistent generalization of MCM}, incorporating its prior and learning to refine it for finer multi-modal UQ. 

\subsection{Training procedure}
\label{sec:training}

\ours accommodates both image-caption and image-label tasks during training and inference.

\vspace{1mm}
\textbf{1) Image-Label Tasks:}  
{Image-label classification} considers a predefined set of $K$ target categories, leading to a \textit{batch-independent} predictive pipeline. Here, the textual representations of categories are obtained using \textit{text templates} (\eg, {"\texttt{A photo of a [CLASS]}"}), resulting in a fixed set of textual captions $\{\bm{t}_j\}_{j \in \{1,...,K\}}$. The predicted concept for an image is then determined as:  
\begin{equation}
\bm{\hat{t}}_i = \argmax_{j\in\{1,...,K\}} p(\bm{t}_{j}|\bm{x}_i).
\end{equation}

This setting is batch-independent, making it suitable for standard classification datasets with predefined labels.

\textbf{2) Image-Caption Tasks:} When a set of textual descriptions is available from an open-vocabulary domain, the goal is to assign the most similar caption to a given input image. It is typically validated using batches of paired images and text descriptions. Given a batch $\mathcal{B}$ of images with associated text descriptions, $\{(\bm{x}_i, \bm{t}_i)\}_{i \in \mathcal{B}}$, the predicted concept for an image is determined as:  
\begin{equation}
\bm{\hat{t}}_i = \argmax_{j\in\mathcal{B}} p(\bm{t}_{j}|\bm{x}_i).
\end{equation}
Here, the performance is \textit{batch-dependent}, with larger batch sizes increasing task complexity.


\vspace{2mm}
\mypar{Training objective.} Our uncertainty module $u_{\theta}$ is trained as a binary classifier to predict VLM zero-shot misclassifications. The parameters $\theta = \{ \theta_{\text{\tiny{XA}}}, \theta_{\text{\tiny{MLP}}}\}$ of our uncertainty model 
are trained by mini-batch gradient descent to minimize the following weighted binary cross-entropy loss:
\begin{equation}
    \mathcal{L}_{\text{wBCE}} = 
    - \frac{1}{B} \sum\limits_{i} \textstyle \big[ w y_i \log \hat y_i + (1 - y_i) \log (1 - \hat y_i) \big]
    \label{eq:opt}
\end{equation}
\noindent where  $y_i = \mathds{1}_{\{ \hat{\bm{t}}_i \neq \bm{t}_i \}}$ is the target label, and $w$ is a weighting factor that mitigates the potential class imbalance between correctly and incorrectly classified examples. This weight is dynamically adjusted based on the empirical classification accuracy of the VLM within each mini-batch:
\begin{equation}
\label{weighting}
    w = \log\Big(1+ \frac{\sum_{i=1}^{B} (1 - y_i)}{\sum_{i=1}^{B} y_i} \Big)
\end{equation}

\noindent Previous UQ methods predict the model’s training loss \cite{confidnet,Learning_loss_AL,pretrained_visual_uncertainties}, assuming a high loss value indicates misclassification. Applying this approach to VLMs would require predicting CLIP’s \cite{clip} contrastive loss or SigLIP’s \cite{siglip} sigmoid loss, which is impractical in a post-hoc setting where the pre-training loss is unknown. In contrast, \ours is loss-agnostic, relying only on whether a prediction is correct. As shown in the experiments \cref{tab:ablation_loss}, training it as a binary classifier consistently outperforms loss prediction for VLMs, naturally aligning with standard error detection metrics.

\begin{table*}[!t]
\scriptsize
\centering
\small
\setlength{\tabcolsep}{4pt}
\resizebox{0.9\linewidth}{!}{
    \begin{tabular}{ c }
    \begin{tabular}{l |cc  cc cc cc cc cc cc}
        \toprule
             \multicolumn{1}{c}{}& \multicolumn{2}{c}{CIFAR-10} & \multicolumn{2}{c }{CIFAR-100} & \multicolumn{2}{c }{Caltech101} & \multicolumn{2}{c}{Flowers102} & \multicolumn{2}{c}{OxfordPets} & \multicolumn{2}{c}{Food101} &  \multicolumn{2}{c}{ImageNet-1k} \\
            \multicolumn{1}{l}{} & \multicolumn{2}{c}{\footnotesize{88.3$\%$}} & \multicolumn{2}{c}{\footnotesize{68.6$\%$}} & \multicolumn{2}{c}{\footnotesize{91.4$\%$}} & \multicolumn{2}{c}{\footnotesize{64.0$\%$}} & \multicolumn{2}{c}{\footnotesize{85.1$\%$}} & \multicolumn{2}{c}{\footnotesize{78.9$\%$}} &  \multicolumn{2}{c}{\footnotesize{62.0$\%$}} \\
             \cmidrule(lr){2-3} \cmidrule(lr){4-5} \cmidrule(lr){6-7} \cmidrule(lr){8-9} \cmidrule(lr){10-11} \cmidrule(lr){12-13} \cmidrule(lr){14-15}
            \multicolumn{1}{c}{} & {\scriptsize AUC$\uparrow$} & {\scriptsize FPR95$\downarrow$} & {\scriptsize AUC$\uparrow$} & {\scriptsize FPR95$\downarrow$} & {\scriptsize AUC$\uparrow$} & {\scriptsize FPR95$\downarrow$} & {\scriptsize AUC $\uparrow$} & {\scriptsize FPR95$\downarrow$} & {\scriptsize AUC $\uparrow$} & {\scriptsize FPR95$\downarrow$}& {\scriptsize AUC $\uparrow$} & {\scriptsize FPR95$\downarrow$}& {\scriptsize AUC $\uparrow$} & {\scriptsize FPR95$\downarrow$}\\
        \midrule 
            MCM \cite{MCM}                    & 89.9 & 52.1 & 82.7 & 67.3 & 88.1 & 68.7 & 86.6 & 68.0 & 87.2 & 59.9 & 86.4 & 63.3 & 80.8 & 71.3  \\
           TS \cite{tscale} + MCM \cite{MCM}  & 89.9 & 51.5 & 83.9 & 68.4 & 90.4 & 55.7 & 86.9 & 66.0 & 89.1 & \underline{55.5} & 86.9 & 62.8 & 80.7 & 71.5  \\
            Entropy \cite{predictive_entropy} & 88.7 & 59.9 & 79.8 & 71.9 & 86.1 & 78.8 & 85.5 & 65.0 & 88.0 & 60.0 & 86.1 & 65.0 & 78.3 & 76.8  \\
            Doctor \cite{doctor}              & 89.5 & 56.5 & 82.3 & 69.7 & 88.7 & 66.5 & 86.7 & 63.9 & 88.9 & 56.6 & 86.8 & 63.4 & 80.3 & 72.9  \\
            Rel-U \cite{relu}                 & 86.2 & 54.4 & 81.0 & 68.2 & 90.2 & 58.5 & 90.0 & 47.3 & 83.5 & 59.3 & 81.8 & 73.4 & 75.1 & 85.0  \\
           LVU~\cite{confidnet, Learning_loss_AL, pretrained_visual_uncertainties}       & \underline{96.6} & \underline{21.2} & 80.3 & 68.5 & 89.8 & 50.9 & \underline{90.5} & \underline{38.3} & 84.1 & 55.7 & 82.7 & 69.9 & 78.7 & 77.0  \\
            BayesVLM \cite{BayesVLM}    & 92.6 & 44.9 & \underline{87.0} & \underline{60.3} & \underline{94.0} & \underline{37.4} & 87.3 & 62.4 & \underline{89.5} & 60.3 & \underline{87.8} & \underline{60.3} & \underline{81.5} & \underline{70.3}  \\
            \rowcolor{ours_color} \textbf{\ours (\textit{Ours})} &  \textbf{98.3} & \textbf{7.7} &  \textbf{91.5} &  \textbf{35.4} &  \textbf{96.7} &  \textbf{18.2} &  \textbf{98.7} &  \textbf{5.1} &  \textbf{94.4} &  \textbf{24.5} &  \textbf{94.8} &  \textbf{28.5} & \textbf{89.5} &  \textbf{47.4} \\
        \end{tabular} \\ 
        \midrule
        \midrule
        \begin{tabular}{l |cc cc  cc cc cc cc| cc}
             \multicolumn{1}{c}{}& \multicolumn{2}{c}{FGVCAircraft} & \multicolumn{2}{c}{EuroSAT} & \multicolumn{2}{c}{StanfordCars} & \multicolumn{2}{c}{DTD} & \multicolumn{2}{c}{SUN397} & \multicolumn{2}{c}{UCF101} & \multicolumn{2}{|c}{\textbf{Average}}\\
             \multicolumn{1}{l}{} & \multicolumn{2}{c}{\footnotesize{18.1$\%$}} & \multicolumn{2}{c}{\footnotesize{35.8$\%$}} & \multicolumn{2}{c}{\footnotesize{60.1$\%$}} & \multicolumn{2}{c}{\footnotesize{43.0$\%$}} & \multicolumn{2}{c}{\footnotesize{62.1$\%$}} & \multicolumn{2}{c}{\footnotesize{61.6$\%$}} &  \multicolumn{2}{c}{\footnotesize{62.7$\%$}} \\
             \cmidrule(lr){2-3} \cmidrule(lr){4-5} \cmidrule(lr){6-7} \cmidrule(lr){8-9} \cmidrule(lr){10-11} \cmidrule(lr){12-13} \cmidrule(lr){14-15}
            \multicolumn{1}{c}{} & {\scriptsize AUC$\uparrow$} & {\scriptsize FPR95$\downarrow$} & {\scriptsize AUC$\uparrow$} & {\scriptsize FPR95$\downarrow$} & {\scriptsize AUC$\uparrow$} & {\scriptsize FPR95$\downarrow$} & {\scriptsize AUC $\uparrow$} & {\scriptsize FPR95$\downarrow$} & {\scriptsize AUC $\uparrow$} & {\scriptsize FPR95$\downarrow$} & {\scriptsize AUC $\uparrow$} & {\scriptsize FPR95$\downarrow$} & {\scriptsize AUC $\uparrow$} & {\scriptsize FPR95$\downarrow$}\\
        \midrule 
            MCM \cite{MCM}                    & \underline{75.7} & 82.9 & 64.1 & 87.6 & 81.4 & 73.4 & 77.4 & 77.9 & 78.8 & 75.9 & 84.1 & 68.9 & 81.8 & 70.6 \\
            TS \cite{tscale} + MCM \cite{MCM} & 74.9 & 82.9 & 63.0 & 88.1 & 81.6 & 71.9 & 76.9 & 78.3 & 79.0 & 75.5 & 84.5 & 69.7 & 82.0 & 69.5 \\
            Entropy \cite{predictive_entropy} & 74.1 & 83.6 & 61.0 & 92.1 & 79.4 & 77.5 & 76.4 & 80.2 & 75.8 & 78.6 & 83.5 & 72.9 & 80.2 & 74.0 \\
            Doctor \cite{doctor}              & 74.8 & 82.9 & 62.4 & 89.5 & 80.9 & 73.1 & 76.9 & 82.6 & 78.2 & 77.2 & 84.5 & 70.2 & 81.6 & 71.2 \\
            Rel-U \cite{relu}                 & 68.6 & \underline{82.5} & 76.3 & 72.1 & 75.5 & 78.9 & 81.4 & 69.7 & 75.2 & 81.3 & 84.1 & 61.1 & 80.7 & 68.6  \\
            LVU~\cite{confidnet, Learning_loss_AL, pretrained_visual_uncertainties}        & 74.8 & 83.5 & \underline{95.9} & \underline{19.3} & 78.4 & 75.4 & \underline{87.2} & \underline{55.9} & 76.7 & 76.9 & \underline{88.6} & \underline{53.6} & \underline{85.0} & \underline{57.4} \\
            BayesVLM \cite{BayesVLM}          & 70.9 & 84.3 & 74.3 & 86.6 & \underline{87.7}  & \underline{63.4} & 77.6 & 77.0 & \underline{80.3} & \underline{73.4} & 84.6 & 66.2 & 84.2 & 65.1\\
           \rowcolor{ours_color} \textbf{\ours (\textit{Ours})} &  \textbf{81.0} & \textbf{71.7} & \textbf{98.8} & \textbf{4.4} &  \textbf{90.1} &  \textbf{46.8} &  \textbf{93.8} &  \textbf{28.8} &  \textbf{88.6} &  \textbf{50.3} &  \textbf{95.9} &  \textbf{20.9} &  \textbf{93.2} &  \textbf{29.9}\\
        \bottomrule
        \end{tabular}
    \end{tabular}
}
\caption{\textbf{Misclassification detection on image-label datasets}. The evaluation leverages each dataset's labeled classes as textual queries, which are fixed for each batch. Values below dataset names denote CLIP zero-shot accuracy on the respective dataset.}
\label{tab:classif_dst}
\end{table*}

\section{Experiments}
\label{sec:experiments}

This section presents experimental results validating our multi-modal uncertainty model. We first focus on predicting VLM failures in zero-shot classification across two setups: \textbf{\textit{i)}} standard image-label classification datasets and \textbf{\textit{ii)}} large-scale image-caption datasets. Next, we experimentally analyze the different components of our model and training procedure, conducting various assessments to evaluate its behavior. We provide qualitative results and visualizations illustrating the ability of \ours~to quantify uncertainty. 

\subsection{Experimental setup}

\mypar{Datasets.} We conduct a benchmark across 16 commonly used datasets to evaluate the ability of VLMs to detect misclassification in image classification. As outlined in \cref{sec:training}, we consider two settings: \textbf{\textit{i)}} image-label datasets and \textbf{\textit{ii)}} image-caption datasets. Image-label datasets are standard datasets used in CLIP's transfer learning~\cite{conditional_prompt_learning,gallop}, covering general object recognition, fine-grained classification, and specialized domains (see \cref{appendix_datasets}). 
~We use official or CoOp~\cite{coop} data splits for evaluation. Image-caption datasets, including CC3M \cite{cc3m}, CC12M~\cite{cc12m}, and LAION-400M~\cite{laion400m}, contain free-text descriptions for each sample. From these datasets, we randomly hold out 1\% of the data for testing. 

\mypar{Implementation details.} We use CLIP ViT-B/32 as the default backbone in all experiments. Full implementation details, along with additional results using other backbones, are provided in Appendix~\ref{appendix_implem_details} and ~\ref{sec:supp-backbones}, respectively.

\vspace{1mm}
\mypar{Baselines.} We explore relevant baselines for uncertainty estimation in vision, as well as recent approaches for VLMs. Further details on these baselines and implementation specifics can be found in~\cref{appendix_baselines}. \textbf{1) Measures of output distribution:} MCM~\cite{MCM}, described in \cref{sec:background}, along with its calibrated variant~\cite{tscale}, provide robust uncertainty estimates without additional training. Similarly, Entropy~\cite{predictive_entropy} and Doctor~\cite{doctor} are commonly used as baselines for uncertainty estimation. \textbf{2) Data-driven predictors:} We implement the recent Rel-U~\cite{relu}, which incorporates cross-label uncertainties in the logit space. However, since Rel-U relies on label-based information, it does not apply to image-caption datasets. Additionally, we assess methods that leverage embedding representations to learn patterns related to uncertainty. Specifically, we compare against baselines that predict classifier loss, such as ConfidNet~\cite{confidnet} and other vision-only estimators~\cite{Learning_loss_AL, pretrained_visual_uncertainties}, collectively referred to as ‘Learning Visual Uncertainty’ (LVU). Finally, we evaluate the most recent post-hoc UQ probabilistic modeling approach designed for VLMs, BayesVLM~\cite{BayesVLM}.

\vspace{1mm}
\mypar{Metrics.} To assess the performance of our uncertainty model, we rely on two standard metrics: \textbf{1)} False Positive Rate at $95\%$ True Positive Rate (\textbf{FPR95}) and \textbf{2)} Area Under the receiver-operating characteristic Curve (\textbf{AUC}). These are commonly used in failure prediction to quantify the model's ability to detect incorrect classifications \cite{confidnet,doctor,relu}.

\subsection{Main results}
\vspace{1mm}
\mypar{Image classification datasets.} In \cref{tab:classif_dst}, we evaluate misclassification detection on CLIP's zero-shot predictions across 13 standard image-label datasets. Notably, MCM emerges as a strong baseline in this setting, even outperforming more complex methods such as Doctor and Rel-U. This advantage arises because these methods rely on task-specific vision classifiers, which are generally less well-calibrated than large-scale pre-trained models.
Consequently, MCM achieves superior performance without requiring post-processing steps, such as temperature scaling~\cite{tscale}, for failure prediction. On the other hand, expressive data-driven methods such as LVU, BayesVLM, and our proposed \ours enable more effective failure detection. 
Our proposed \ours ranks first across all datasets and metrics. Its novel architectural design leverages the class-semantic information embedded in MCM while also capturing uncertainty patterns specific to each sample's ambiguities. As a clear demonstration of its effectiveness, \ours achieves remarkable improvements in FPR95, surpassing MCM, BayesVLM, and LVU by margins of $-40.7$, $-35.2$, and $-27.5$, respectively (\cref{tab:classif_dst}, Average column). 

CLIP’s zero-shot accuracy for each dataset is reported in \cref{tab:classif_dst}. MCM and BayesVLM performances closely follow CLIP’s accuracy, struggling in low zero-shot accuracy settings: they achieve only $64.1\%$ and $74.3\%$ AUC in EuroSAT, a dataset on which CLIP’s accuracy is $35.8\%$. This limitation makes them unreliable when zero-shot accuracy is low -- an unpredictable scenario in real-world settings. In contrast, \ours remains effective across all accuracy regimes. See~\cref{sec:supp-add} for further analysis.

\vspace{1mm}
\noindent\textbf{Large-scale image-caption datasets.} We now evaluate the ability of state-of-the-art methods to detect failures in open-vocabulary settings, where tasks vary based on the inference batch and its specific captions. \cref{tab:captions_dst} presents results for applicable baselines in this challenging setting, along with the proposed \ours. Unlike in previous experiments on image-label datasets, LVU fails to outperform even the MCM baseline, underscoring the importance of considering target objectives alongside sample-related uncertainties in open-vocabulary scenarios. As a result, methods specifically designed for vision-language models, such as BayesVLM, achieve superior performance across all three evaluated datasets. Our proposed \ours achieves the best performance, consistently surpassing BayesVLM with FPR95 improvements of $-21.5$, $-28.1$, and $-5.2$ on CC3M, CC12M, and LAION-400M, respectively. This advantage stems from our explicit modeling of misclassification errors, whereas BayesVLM focuses on the uncertainty of similarities between individual embeddings, which is more implicit.

\begin{table}[!t]
    \centering
    \small
    \setlength{\tabcolsep}{3pt}
    \resizebox{0.85\linewidth}{!}{
    \begin{tabular}{l|cc cc cc}
    \toprule
         \multicolumn{1}{c}{}& \multicolumn{2}{c }{CC3M} & \multicolumn{2}{c }{CC12M} & \multicolumn{2}{c}{LAION-400M} \\
         \multicolumn{1}{c}{} & \multicolumn{2}{c}{\footnotesize{58.8$\%$}} & \multicolumn{2}{c}{\footnotesize{73.5$\%$}} & \multicolumn{2}{c}{\footnotesize{90.5$\%$}} \\
         \cmidrule(lr){2-3} \cmidrule(lr){4-5} \cmidrule(lr){6-7} 
        \multicolumn{1}{c}{} & {\scriptsize AUC$\uparrow$} & {\scriptsize FPR95$\downarrow$} & {\scriptsize AUC$\uparrow$} & {\scriptsize FPR95$\downarrow$} & {\scriptsize AUC$\uparrow$} & {\scriptsize FPR95$\downarrow$} \\
    \midrule 
        MCM \cite{MCM} & 83.9 & 69.0 & 88.8 & 58.8 & 91.7 & 50.2 \\
        Entropy \cite{predictive_entropy}& 82.5 & 73.3 & 87.7 & 63.0 & 89.4 & 62.5  \\
        Doctor \cite{doctor} & 83.7 & 70.1 & 88.6 & 59.9 & 91.2 & 54.5   \\
        LVU~\cite{confidnet, Learning_loss_AL, pretrained_visual_uncertainties} & 69.3 & 82.5 & 74.4 & 76.5 & 80.2 & 72.3 \\
        BayesVLM \cite{BayesVLM} & \underline{87.1} & \underline{62.6} & \underline{90.9} & \underline{53.3} & \underline{95.1} & \underline{26.4} \\
        \rowcolor{ours_color} \textbf{\ours (\textit{Ours})} & \textbf{91.4}& \textbf{41.1} & \textbf{95.2} & \textbf{25.2} & \textbf{97.3} & \textbf{21.2}\\
    \bottomrule
    \end{tabular}
    }
    \caption{\textbf{Misclassification detection on image-caption datasets}. The evaluation uses the captions of each randomly-retrieved batch as textual queries. Hence, the textual queries vary for each batch. Results reported with a batch size of 1024 samples for inference.}
    \label{tab:captions_dst}
\end{table}

\subsection{Ablation studies} 

\begin{figure}[t]
\centering
\scalebox{0.75}{
    \begin{subfigure}{0.48\linewidth}
        \centering
        \includegraphics[width=\linewidth, height=0.85\linewidth]{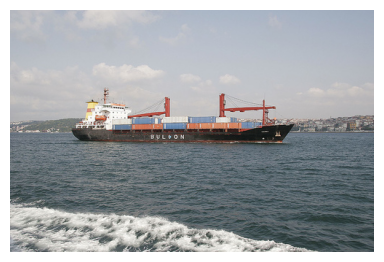}
        \subcaption*{
            \centering
            \small{\makecell{\textbf{GT}: \textit{"container ship"} \\ \textbf{Pred}: \textit{"ocean liner"}}
            }
        }
    \end{subfigure}
    \hfill
    \begin{subfigure}{0.48\linewidth}
        \centering
        \includegraphics[width=\linewidth, height=0.85\linewidth]{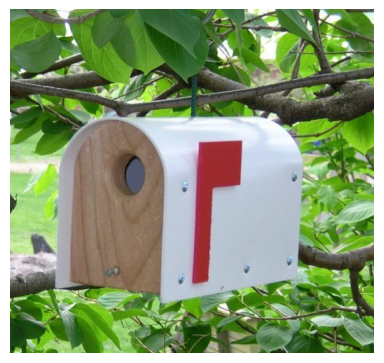}
        \subcaption*{
            \centering
            \small{\makecell{\textbf{GT}: \textit{"mailbox"} \\ \textbf{Pred}: \textit{"birdhouse"}}
            }
        }
    \end{subfigure}}
    \caption{\textbf{Two qualitative examples.} Misclassifications detected by \ours, but not MCM \cite{MCM} and visual-only baselines \cite{pretrained_visual_uncertainties,confidnet}.} 
    \label{fig:failure_quali_failure_pred}
\end{figure}

\mypar{Architectural design of \ours.} \cref{tab:ablation_input} highlights the contributions of different components in our model, namely the cross-attention mechanism and the inclusion of the predicted caption as input. As observed in~\cite{confidnet, pretrained_visual_uncertainties}, using \textbf{only visual information} (first row) is effective on CIFAR-10 but struggles on larger datasets with fine-grained or semantically similar concepts. For instance, on ImageNet-1k, visual information alone fails to outperform MCM. Incorporating the \textbf{predicted class textual embedding} (second row) significantly improves performance across all datasets, yielding a $+10$ AUC gain on ImageNet-1k and $+14$ AUC on CC12M. This additional input helps the model to handle class ambiguity, which is crucial for datasets where categories are easily confused. For example, as shown in \cref{fig:failure_quali_failure_pred}, the class \texttt{container ship} is often misclassified as \texttt{ocean liner}, another type of boat. Finally, the \textbf{cross-attention module} (third row) enables the model to integrate contextual information from all candidate classes or captions. By re-contextualizing predictions among available textual inputs, this mechanism proves particularly beneficial for CC12M, where captions change dynamically across batches. Unlike CIFAR-10 and ImageNet-1k, where class sets are fixed, omitting cross-attention in CC12M results in an AUC plateau at 88.9, only slightly above MCM. Incorporating this mechanism enhances the model's ability to detect ambiguities and assess potential errors.

\begin{table}[!h]
    \centering
    \small
    \setlength{\tabcolsep}{2pt}
    \resizebox{0.9\linewidth}{!}{
    \begin{tabular}{c c c  cc cc cc}
    \toprule
          \multirow{2}{*}{\makecell{Visual \\ embed.}} & \multirow{2}{*}{\makecell{Cross\\attention}} & \multirow{2}{*}{\makecell{Predicted \\ caption}} & \multicolumn{2}{c}{CIFAR-10} & \multicolumn{2}{c }{ImageNet-1k} & \multicolumn{2}{c}{CC12M} \\
         \cmidrule(lr){4-5} \cmidrule(lr){6-7} \cmidrule(lr){8-9} 
         & & &  {\scriptsize AUC$\uparrow$} & {\scriptsize FPR95$\downarrow$} & {\scriptsize AUC$\uparrow$} &  {\scriptsize FPR95$\downarrow$} & {\scriptsize AUC$\uparrow$} & {\scriptsize FPR95$\downarrow$} \\
        \midrule 
        \cmark & \xmark & \xmark & 96.4  &  21.8 & 78.7  &77.0 & 74.0 &  76.5   \\
         \cmark & \xmark & \cmark & 97.9  & 10.8  & 88.8 &50.1  & 88.9 & 48.9  \\
       \cmark & \cmark & \xmark &  97.7 & 11.4  &86.1  & 63.5  & 93.6 & 37.0\\
        \cmark & \cmark & \cmark & \textbf{98.3} & \textbf{8.2}& \textbf{89.5} & \textbf{47.4}& \textbf{95.2} & \textbf{25.2}  \\
        \midrule
        \midrule
        \multicolumn{3}{c}{MCM \cite{MCM}} & 89.9 & 52.1 & 80.8 & 71.3 & 88.8 & 58.8 \\
    \bottomrule
    \end{tabular}
    }
    \caption{\textbf{Ablation on different components of \ours}.}
    \label{tab:ablation_input}
\end{table}

\vspace{1mm}
\mypar{Loss function design for failure prediction.} In \cref{sec:training}, we discussed our optimization target and choice of loss function. Unlike prior works~\cite{confidnet,Learning_loss_AL, pretrained_visual_uncertainties}, which treat the problem as a regression task by directly predicting the test-time loss and optimizing it with MSE, we instead frame it as a binary classification task, distinguishing between errors and correct predictions. As shown in \cref{tab:ablation_loss}, this approach consistently outperforms MSE-based loss approximations, yielding a $+3$ AUC improvement on ImageNet-1k while reducing FPR95 by 16 points. Additionally, our automatic weighting of the two classes (misclassified and correctly classified) further enhances the performance, resulting in a $+1$ AUC gain on ImageNet-1k and a 2-point reduction in FPR95 on CIFAR-10.

\begin{table}[!hbt]
    \centering
    \small
    \setlength{\tabcolsep}{4pt}
    \resizebox{0.8\linewidth}{!}{
    \begin{tabular}{l  cc  cc }
    \toprule
          \multicolumn{1}{c}{} & \multicolumn{2}{c}{CIFAR-10} & \multicolumn{2}{c}{ImageNet-1k}  \\
         \cmidrule(lr){2-3} \cmidrule(lr){4-5} 
         \multicolumn{1}{c}{} &  {\scriptsize AUC$\uparrow$} & {\scriptsize FPR95$\downarrow$} & {\scriptsize AUC$\uparrow$} &  {\scriptsize FPR95$\downarrow$}  \\
        \midrule 
        MSE \cite{confidnet,pretrained_visual_uncertainties} & 95.1  & 30.4   & 84.8  &  64.4  \\
        \hspace{2mm}w/ Weighting & 95.6   & 29.6  & 85.7  &  63.2   \\ \hdashline
        BCE  &  97.7  &   10.5 &  88.6 & 48.4   \\
        \hspace{2mm}\textbf{w/ Weighting (\textit{Ours})} & \textbf{98.3} & \textbf{8.2}  & \textbf{89.5}& \textbf{47.4}   \\
    \bottomrule
    \end{tabular}
    }
    \caption{\textbf{Role of the proposed weighted loss}. Effect of loss function choice and adaptive weighting in \cref{weighting} for failure detection.}
    \label{tab:ablation_loss}
\end{table}

\vspace{-0.5em}
\subsection{In-depth analysis}

\begin{figure}[t]
\centering
    \begin{subfigure}{0.49\linewidth}
        \centering
         \includegraphics[width=.99\linewidth]{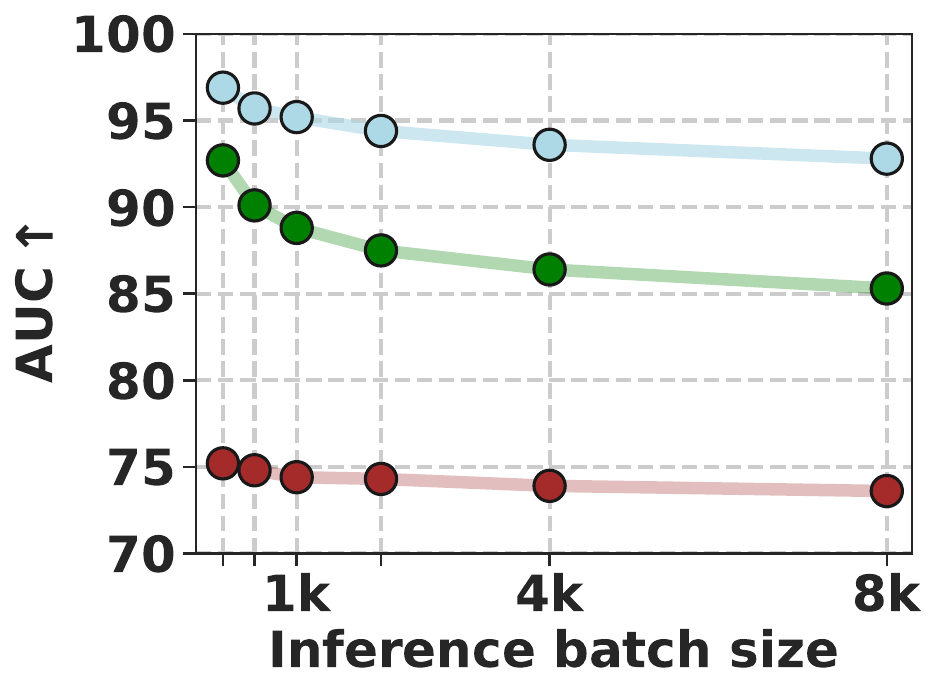} 
        \subcaption*{AUC}
    \end{subfigure}
    \hfill
    \begin{subfigure}{0.49\linewidth}
        \centering
\includegraphics[width=.99\linewidth]{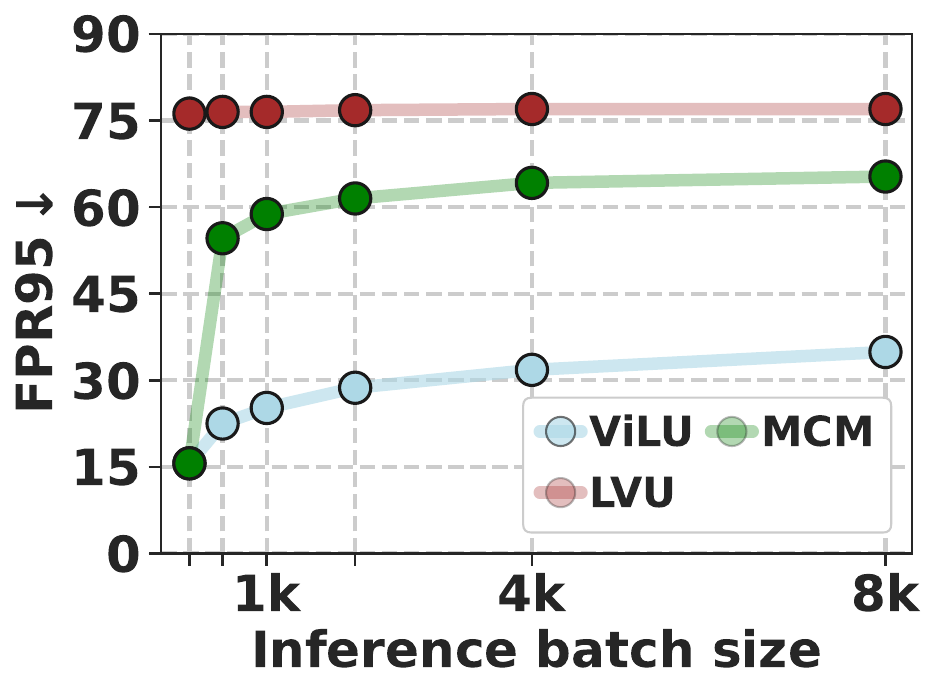} \\
        \subcaption*{FPR95}
    \end{subfigure}
\vspace{-0.5em}
\caption{\textbf{Robustness to image-caption task complexity.} Inference batch size effect in failure detection for \ours (CC12M).}
        \label{fig:batch_sizes}
\end{figure}

\mypar{Classification complexity on image-caption datasets.} \cref{fig:batch_sizes} explores the performance of \ours in challenging image-caption classification tasks, where difficulty is determined by the number of concepts to be distinguished simultaneously, \ie, batch size used for inference. Naturally, increasing the batch size makes the classification more complex, making errors harder to detect for methods that rely on semantic relationships among query concepts, such as \ours or MCM. However, \ours consistently outperforms MCM across all batch sizes. Notably, this configuration does not affect prior LVU methods, which estimate vision-only uncertainty. However, they fall short in performance, particularly compared to the proposed \ours.

\begin{figure}[h!]
    \begin{center}
        \setlength{\tabcolsep}{1pt}
        \begin{tabular}{cc}

         \hspace{-3mm}
         \includegraphics[width=.5\linewidth]{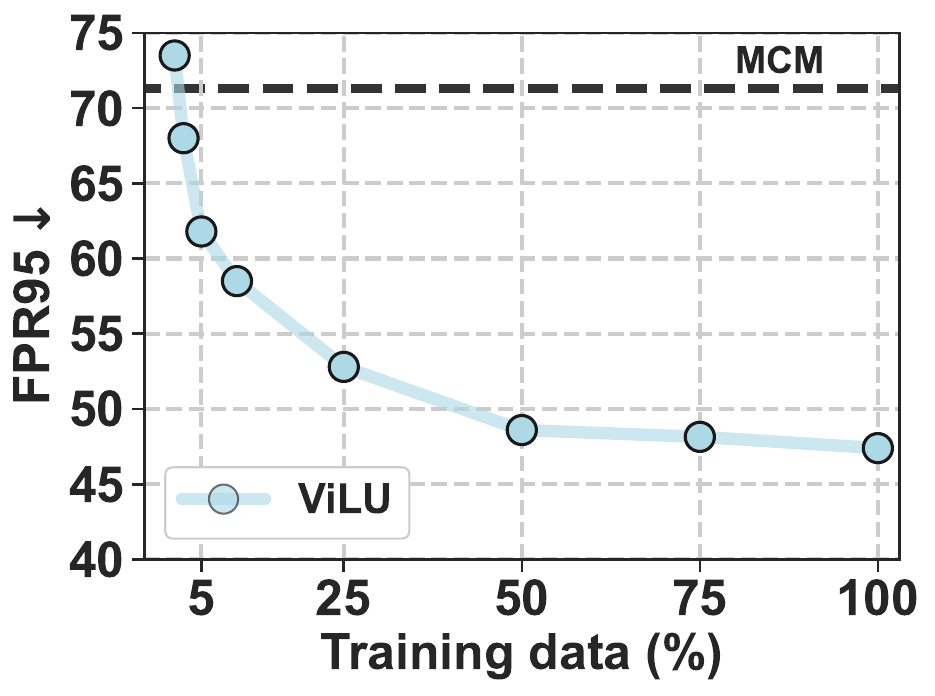} 
         & \includegraphics[width=.5\linewidth]{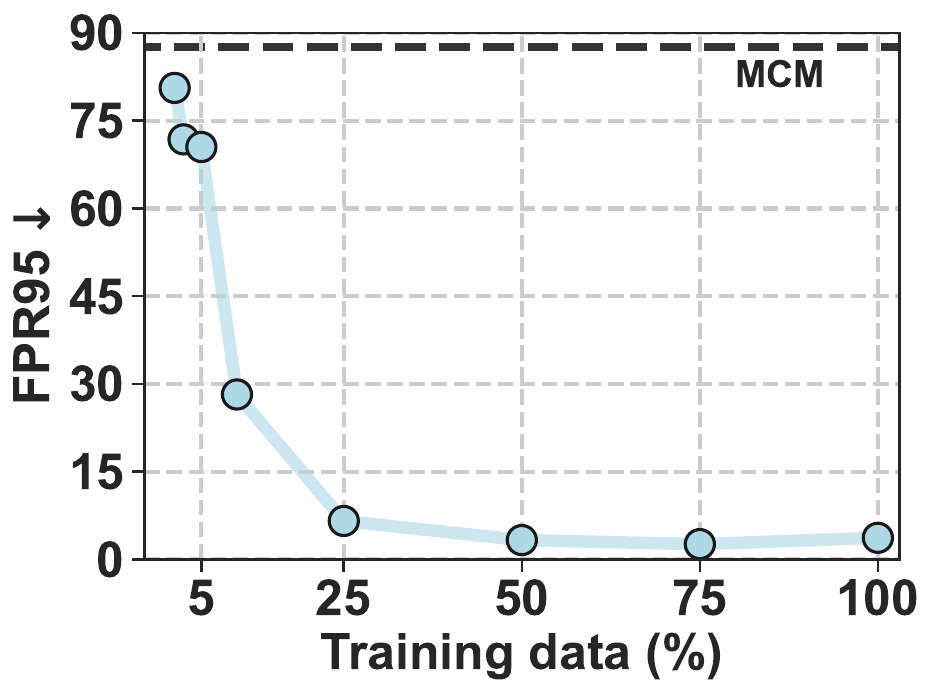} \\

          \multicolumn{1}{c}{\small{\textbf{(a) ImageNet-1k}}} & \multicolumn{1}{c}{\small{
          \textbf{(b) EuroSAT}}} \\
        \end{tabular}
        \caption{\textbf{Data-efficiency.} Performance, in terms of FPR95$\downarrow$, w.r.t. the available data for training \ours.}
        \label{fig:data_ratio}
    \end{center}
\end{figure}

\vspace{2mm}
\mypar{Requirements on training data.} As a data-driven uncertainty quantifier, the proposed \ours requires a subset of image-caption or image-label examples. \cref{fig:data_ratio} illustrates the data-efficiency of the proposed approach for uncertainty quantification. The results showcase the efficiency of \ours, which \textit{requires only a small amount of data to surpass the strong baseline MCM}, \eg, $2.5\%$ for ImageNet and even less for specialized datasets such as EuroSAT.

\vspace{-1em}
\noindent \textbf{Cross-datasets generalization.} To evaluate ViLU’s robustness to dataset shifts and its ability to generalize without per-dataset tuning, we report in \cref{tab:fpr_comparison} its zero-shot transfer performance when pre-trained on the large-scale CC12M dataset. ViLU consistently outperforms MCM across all 12 datasets in this pure transfer setting. The table also highlights ViLU’s clear advantage over LVU (also trained on CC12M), underscoring the benefit of explicitly modeling vision-language uncertainty for zero-shot generalization. While these results demonstrate strong transfer capabilities, generalization can still be improved. In \cref{supp:sec:fpr_comparison}, we explore how smarter batch sampling strategies aimed at improving concept coverage during pre-training could further enhance performance.

\begin{table}[!ht]
\setlength{\tabcolsep}{6pt}
\centering
\scriptsize
\resizebox{0.65\linewidth}{!}{
    \begin{tabular}{l ccc}
    \hline
    Dataset & MCM & LVU & \textbf{ViLU}\\
    \hline
    CIFAR-10        & \textbf{52.1}   & 77.2 & \ourcell{54.2}           \\
    CIFAR-100       & 67.3    & 83.8 & \ourcell{\textbf{59.9}}         \\
    Caltech101      & 68.7    & 82.5 & \ourcell{\textbf{48.8}} \\
    Flowers102      & 68.0    & 96.8 & \ourcell{\textbf{67.4}}          \\
    OxfordPets      & 59.9    & 93.1 & \ourcell{\textbf{58.1}} \\
    Food101         & \textbf{63.3}    & 87.2 & \ourcell{67.4}           \\
    FGVCAircraft    & 82.9   & 94.5 & \ourcell{\textbf{82.3}}           \\
    EuroSAT         & 87.6    & 88.2 & \ourcell{\textbf{85.7}}       \\
    DTD             & \textbf{77.9}   &  93.1 & \ourcell{78.2}          \\
    SUN397          & 75.9     & 90.1 & \ourcell{\textbf{72.7}} \\
    StanfordCars    & \textbf{73.4}  & 92.6 & \ourcell{84.1}  \\
    UCF101          & 68.9   & 90.4 & \ourcell{\textbf{63.8}} \\
    \hline
    \textbf{Average} & 70.5 & 89.1 & \ourcell{\textbf{68.6}}\\
    \hline
    \end{tabular}
}
\vspace{-0.5em}
\caption{FPR95$\downarrow$ across datasets. Zero-shot performance when pre-trained on CC12M. 
}
\label{tab:fpr_comparison}
\end{table}


\vspace{1mm}
\mypar{Qualitative assessment.} 
In \cref{fig:failure_quali_failure_pred}, we present two qualitative examples from ImageNet where the VLM misclassified the input images. On the left, the model predicted \texttt{ocean liner} instead of \texttt{container ship}, likely due to their shared visual features as large vessels. On the right, a \texttt{mailbox} was misclassified as a \texttt{birdhouse}, possibly influenced by the surrounding vegetation and the mailbox’s opening, which resemble typical birdhouse features. \ours effectively captures both sources of ambiguity: semantic confusion between visually similar classes and misinterpretations driven by contextual cues.

\begin{figure}[!ht]
    \centering
    \hspace{-3mm} \includegraphics[width=\linewidth]{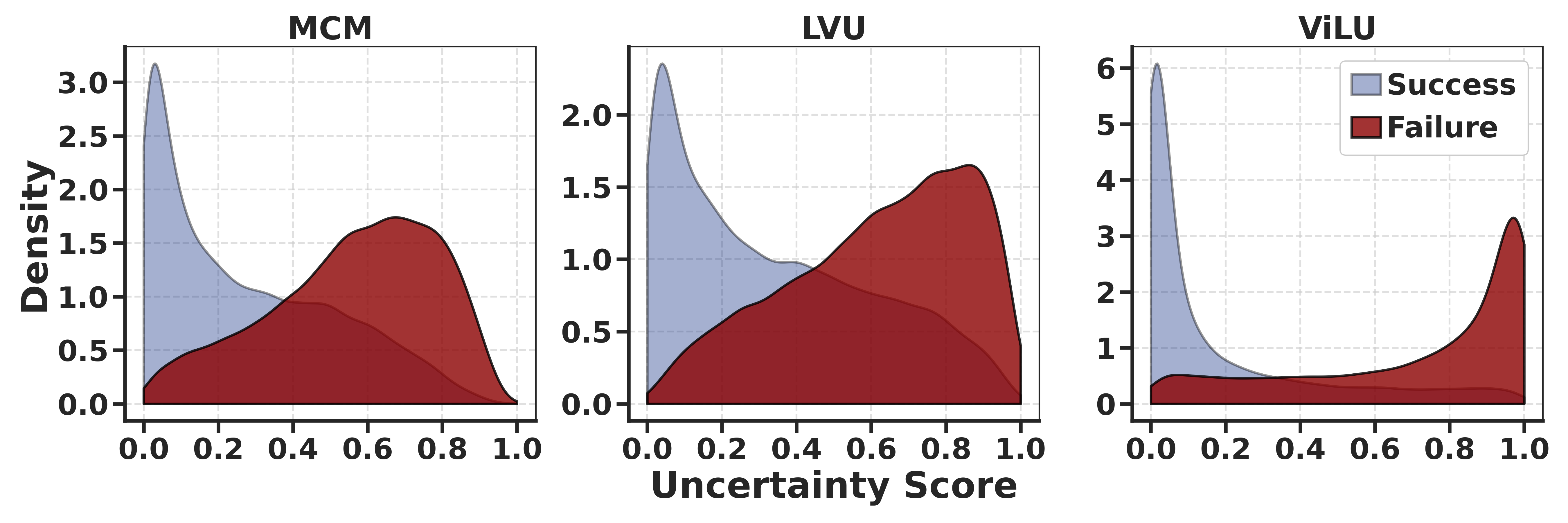}
    \vspace{-0.75em}
    \caption{\textbf{Uncertainty score distribution.} Predictions for correctly and incorrectly classified samples on ImageNet.}
    \label{fig:failure_curve}
\end{figure}

\vspace{1mm}
\mypar{Uncertainty distribution score.}
\cref{fig:failure_curve} illustrates the effectiveness of different uncertainty estimation methods in distinguishing between correctly and incorrectly classified samples on ImageNet. In MCM and LVU, the uncertainty distributions for success and failure overlap significantly. In contrast, \ours produces a more distinct separation. Indeed, misclassified samples receive high uncertainty (peaking near 1.0), while correct predictions have low uncertainty (peaking near 0.0). Its higher density at the extremes indicates a well-calibrated uncertainty measure.

\vspace{-1em}
\section{Conclusion}


We have presented our \ours method, a new uncertainty quantification approach for failure prediction of pre-trained VLMs. \ours learns an appropriate uncertainty embedding space including fine interactions between visual and concepts ambiguities to learn an effective binary failure predictor on the downstream task.  
Extensive experiments on several datasets show the significant gain of our method compared to state-of-the-art UQ methods for failure prediction. In addition, ablation studies  clearly validate our architectural choices and training design. We also show that \ours remains effective even in the low-performance regime of VLMs.
~Future works include adjusting \ours in the context of domain adaptation and test-time adaptation of VLMs.

%
 


\newpage
\section*{Acknowledgment}
This work was supported by the French National Research Agency (Agence Nationale de la Recherche, ANR) under grants from project DIAMELEX (ANR-20-CE45-0026) and project RODEO (ANR-24-CE23-5886). It was granted access to the HPC resources of IDRIS under the allocation AD011013370R2 made by GENCI.

{
    \small
    \bibliographystyle{ieeenat_fullname}
    \bibliography{main}
}

\newpage
\appendix

\setcounter{section}{0}
\renewcommand{\thesection}{\Alph{section}}

\maketitlesupplementary

\section{Additional details on experimental setup}
\subsection{Datasets}
\label{appendix_datasets}
In this section, we provide additional details on the image–label datasets used in the main experiments presented in the paper. These span general object recognition, fine-grained classification, and specialized domains. The datasets include ImageNet-1k~\cite{deng2009imagenet}, CIFAR-10~\cite{cifar100}, CIFAR-100~\cite{cifar100}, SUN397~\cite{sun397}, FGVCAircraft~\cite{aircraft}, EuroSAT~\cite{eurosat}, StanfordCars~\cite{stanfordcars}, Food101~\cite{food101}, OxfordPets~\cite{oxfordpets}, Flowers102~\cite{flowers102}, Caltech101~\cite{caltech}, DTD~\cite{dtd}, and UCF101~\cite{ucf101}.

\subsection{Implementation details}
\label{appendix_implem_details}
As mentioned in the main paper, we use CLIP ViT-B/32 backbone in all our experiments. We provide additional results using other CLIP backbones and architectures, such as SigLIP~\cite{siglip} in \cref{sec:supp-backbones}. Regarding the proposed \ours, the MLP layer for misclassification prediction follows a four-layer architecture with dimensions $\left[512,256,128,1\right]$ and ReLU activations. Training is performed using SGD as the optimizer, with the cross-attention layers remaining frozen during the first epoch. We select the learning rate through a grid search over  $\{10^{-1},10^{-2},10^{-3}\}$ and explore batch sizes among $\{128,256,512,1024\}$. 

\subsection{Baselines \& Implementation}
\label{appendix_baselines}
\begin{itemize}
    \item \textbf{MCM \cite{MCM}:} Maximum Concept Matching (MCM) estimates uncertainty in VLMs by leveraging the softmax probability distribution over all classes or captions. It selects the most likely caption for an image based on the highest probability score, providing a natural measure of confidence in the model’s predictions. No additional training is required since MCM directly uses the model’s output probabilities.
    
    \item \textbf{MCM + TS \cite{tscale}:} This method extends MCM by applying Temperature Scaling (TS) to adjust the softmax probabilities better. TS optimizes the temperature parameter to refine the confidence scores, leading to more calibrated uncertainty estimates. The multiplicative temperature parameter is learned using the whole training dataset to minimize expected calibration error, using LBFGS optimizer.
    
    \item \textbf{Entropy} \cite{predictive_entropy}: This method quantifies uncertainty in neural network predictions by calculating the Shannon entropy of the output probability distribution. High entropy indicates more significant uncertainty, as the model assigns similar probabilities across multiple classes, reflecting ambiguity in its prediction. On the other hand, low entropy signifies confidence, with the model favoring a specific class. Entropy-based uncertainty estimation does not require additional training.
    
    \item \textbf{DOCTOR} \cite{doctor}: DOCTOR quantifies uncertainty by analyzing the confidence distribution of the model’s predictions. It computes the Rényi entropy of order two, a measure based on the squared probabilities assigned to each class, emphasizing how concentrated or dispersed the probability mass is. A prediction with one dominant probability value will yield a low uncertainty score, while a more evenly spread distribution results in higher uncertainty. This method does not require additional training and operates directly on the model’s softmax outputs.
    
    \item \textbf{Rel-U }\cite{relu}:  Rel-U is a data-driven method that incorporates cross-label uncertainties directly in the logit space. Learning relationships between class logits provides a refined estimation of uncertainty beyond traditional confidence scores. Due to its reliance on a cross-label cost penalty matrix, Rel-U does not apply to image-text datasets where labels are absent. Rel-U's hyper-parameters are fixed to $\lambda=0.15$ and $T=0.5$ greedily, since they provided the best performance.
    
    \item \textbf{Learning Visual Uncertainties (LVU) \cite{confidnet,pretrained_visual_uncertainties,Learning_loss_AL}:} LVU refers to a class of models designed to predict the loss of a visual backbone as a means to estimate potential errors. ConfidNet \cite{confidnet} established that accurately predicting uncertainty is equivalent to estimating the model's loss---if a model can predict the loss of its visual backbone, it inherently quantifies its error. Another approach, Pretrained Visual Uncertainties \cite{pretrained_visual_uncertainties}, follows a similar principle by learning to predict backbone loss, leveraging pretraining on ImageNet-21k.\\
    
    \textbf{Implementation:} To evaluate the LVU baseline, we use the same MLP architecture as our model but restrict the input to the visual token only. Additionally, following \cite{confidnet, pretrained_visual_uncertainties}, this baseline is trained with an MSE loss, in contrast to our method, which uses a BCE loss.

    \item \textbf{ProbVLM \cite{probvlm}:} ProbVLM introduces a probabilistic adapter that estimates probability distributions for embeddings of pre-trained VLMs. This is achieved through inter- and intra-modal alignment in a post-hoc manner. The goal is to capture the inherent ambiguity in embeddings, reflecting the fact that multiple samples can represent the same concept in the physical world. This method enhances the calibration of embedding uncertainties in retrieval tasks and benefits downstream applications like active learning and model selection.\\
    \textbf{Implementation:} ProbVLM models probability distributions over the embeddings of image and text modalities. However, it does not explicitly model the uncertainty in their interaction via cosine similarity. As a result, directly adapting the method for image classification is not straightforward. 
    We attempted to include ProbVLM in our baseline comparison by using its proposed visual aleatoric uncertainty metric, but it resulted in nearly random failure prediction performance. Additionally, we explored using its cross-modal loss as an uncertainty logit, applying a softmax transformation, but this approach also proved ineffective.
    In contrast, BayesVLM addresses this limitation by modeling the uncertainty over the similarity computation, enabling a more principled approach to downstream tasks like image classification.

    \item \textbf{BayesVLM \cite{BayesVLM}:} BayesVLM is a training-free method for estimating predictive uncertainty. It employs a post-hoc approximation of the Bayesian posterior, allowing for analytic computation of uncertainty propagation through the VLM. By approximating the Bayesian posterior over model parameters, BayesVLM captures uncertainties inherent to the model itself (image and text encoders). These model uncertainties are then propagated through the VLM to produce uncertainty estimates for predictions.\\
    \textbf{Implementation:} To evaluate BayesVLM, we follow the implementation provided in its official Github repository \url{https://github.com/AaltoML/BayesVLM}
\end{itemize}
\vspace{1em}

\section{Additional details on ViLU}
\label{sec:supp-vilu_details}

\subsection{Bilinear interpretation of MCM}
\label{supp:vilu_bilinear}
In \cref{sec:archi} of the main paper, we mentioned that ViLU is a consistent generalization of MCM. More precisely, the uncertainty module $g_\theta$ can model the unnormalized MCM score by approximating the following bilinear form on $\bm{z}_{\text{\tiny{ViLU}}}=(\bm{z}_v, \bm{z}_{\hat{t}}, \bm{z}^{\alpha}_{t})$:
\begin{align}
     g_\theta(\bm{z}_{\text{\tiny{ViLU}}}) = \frac{1}{2}~\bm{z}_{\text{\tiny{ViLU}}}^T ~A~ \bm{z}_{\text{\tiny{ViLU}}}=\bm{z}_v^{\top} \bm{z}_{\hat{t}},
\end{align}
with $\quad A = \Bigg(
\begin{array}{ccc}
\mathbf{0} & \mathrm{I}_d & \mathbf{0} \\[-0.25ex]
 \mathrm{I}_d & \mathbf{0} & \mathbf{0} \\[-0.25ex]
 \mathbf{0} & \mathbf{0} & \mathbf{0}
\end{array}
\Bigg) \in \mathbb{R}^{3d \times 3d}$.\\

\subsection{ViLU variant for generalization experiments}
\label{supp:vilu_variant}

For the generalization experiments presented in the main paper (cross-dataset transfer) and the supplementary material (domain generalization and concept coverage), we used a slightly modified version of ViLU. Specifically, the MCM score was explicitly provided as an additional input to the uncertainty module $g_\theta$, alongside the visual and textual embeddings. While the original design of ViLU allows $g_\theta$ to model this behavior implicitly through interactions between the modalities, we found that explicitly including the MCM score improves uncertainty generalization.\\

\section{Additional experimental results}
\label{sec:supp-add}
\subsection{Impact of MLP depth on performance} 
\begin{figure}[!h]
    \centering
    \includegraphics[width=\linewidth]{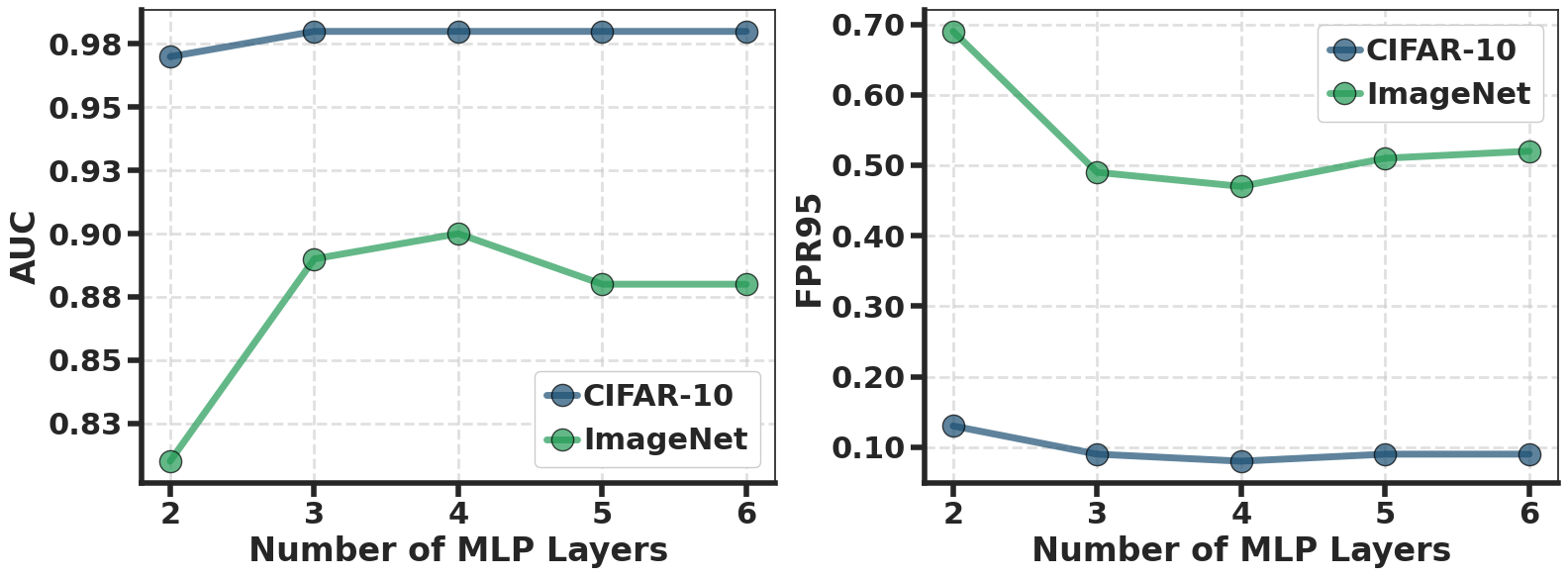}
    \caption{\textbf{Impact of MLP depth.} Performance of \ours on ImageNet and CIFAR-10 for different MLP depths.}
    \label{fig:mlp}
\end{figure}

The results in \cref{fig:mlp} show that \ours is relatively robust to MLP depth variations, particularly on ImageNet, where performance remains stable across different configurations. Across all tested datasets, a depth of 4 layers consistently achieved strong results, suggesting that this architecture provides a good balance between expressiveness and generalization for failure prediction.

\subsection{Robustness to image-text task complexity}
We analyze in \cref{tab:batchsize} how inference-time batch size affects failure detection performance for MCM, LVU, and \ours on the CC12M dataset. As batch size increases, the number of candidate captions used during inference grows, introducing more semantic competition and making the task more complex. Despite this, \ours consistently outperforms both MCM and LVU across all tested settings. Notably, even under very large batch sizes—16,384 and 32,768—\ours maintains strong performance, with only moderate degradation in AUC and FPR95. These results confirm the robustness of our method to increased image-text ambiguity at test time.
\begin{table}[!hbt]
    \centering
    \small
    \resizebox{\linewidth}{!}{
    \begin{tabular}{c|cc cc cc} 
    \toprule
         \multicolumn{1}{c}{}& \multicolumn{2}{c }{MCM \cite{MCM}} & \multicolumn{2}{c }{LVU \cite{confidnet,pretrained_visual_uncertainties,Learning_loss_AL}} & \multicolumn{2}{c}{\ours} \\
         \cmidrule(lr){2-3} \cmidrule(lr){4-5} \cmidrule(lr){6-7} 
        \multicolumn{1}{c}{Batch Size} & {\scriptsize AUC$\uparrow$} & {\scriptsize FPR95$\downarrow$} & {\scriptsize AUC$\uparrow$} & {\scriptsize FPR95$\downarrow$} & {\scriptsize AUC$\uparrow$} & {\scriptsize FPR95$\downarrow$} \\
    \midrule 
        128  & 92.7 & 15.6 & 75.2 & 76.2 & \textbf{96.9} & \textbf{15.6}\\
        512 & 90.1 & 54.6 & 74.8 & 76.5 & \textbf{95.7} & \textbf{22.5}\\ 
        1024 & 88.8 & 58.8 & 74.4 & 76.5 & \textbf{95.2} & \textbf{25.2}\\
        2048 & 87.5 & 61.5 & 74.3 & 76.8 & \textbf{94.4} & \textbf{28.7}\\
        4096 &  86.4 & 64.2 & 73.9 & 77.0 & \textbf{93.6} & \textbf{31.8 }\\
        8192 & 85.3 & 65.3 & 73.6 & 77.0 & \textbf{92.8}  & \textbf{34.9 }\\
        16384 & 84.5 & 66.4 & 73.3 & 76.7 & \textbf{91.9} & \textbf{37.8} \\
        32768 & 83.7 & 67.0 & 73.2 & 76.6 & \textbf{91.1} & \textbf{39.9} \\
    \bottomrule
    \end{tabular}
    }
    \caption{Numerical results corresponding to \cref{fig:batch_sizes}, showing the effect of inference batch size on failure detection for \ours (on CC12M).}
    \label{tab:batchsize}
\end{table}

\subsection{Reliability of misclassification detection}
\begin{figure}[!ht]
    \begin{center}
        \setlength{\tabcolsep}{1pt}
\scalebox{1}{
        \begin{tabular}{cc}
         \hspace{-0.75em}\includegraphics[width=.5\linewidth]{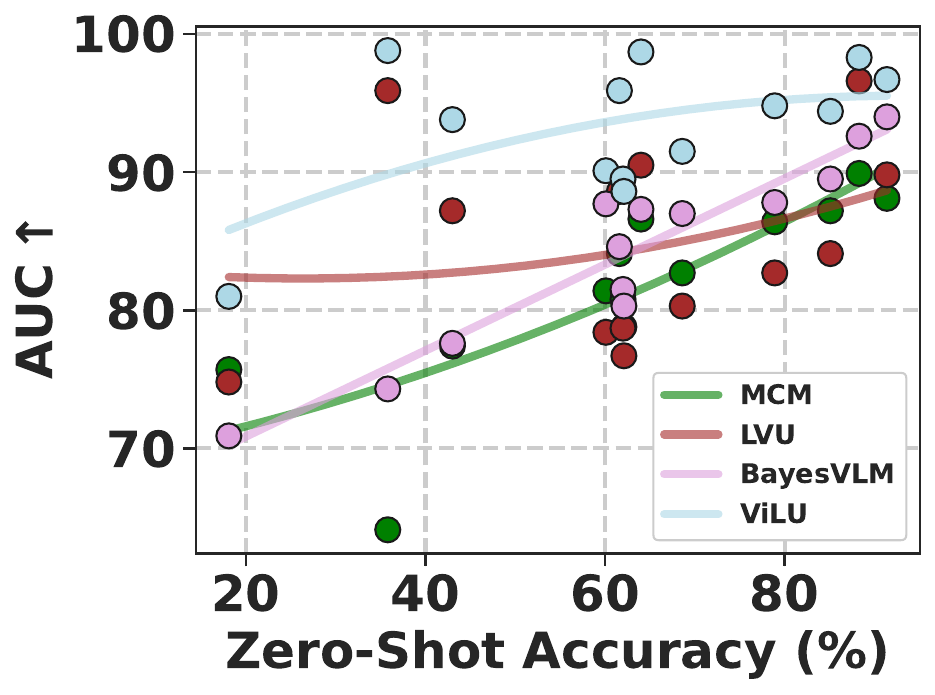} 
         & \includegraphics[width=.5\linewidth]{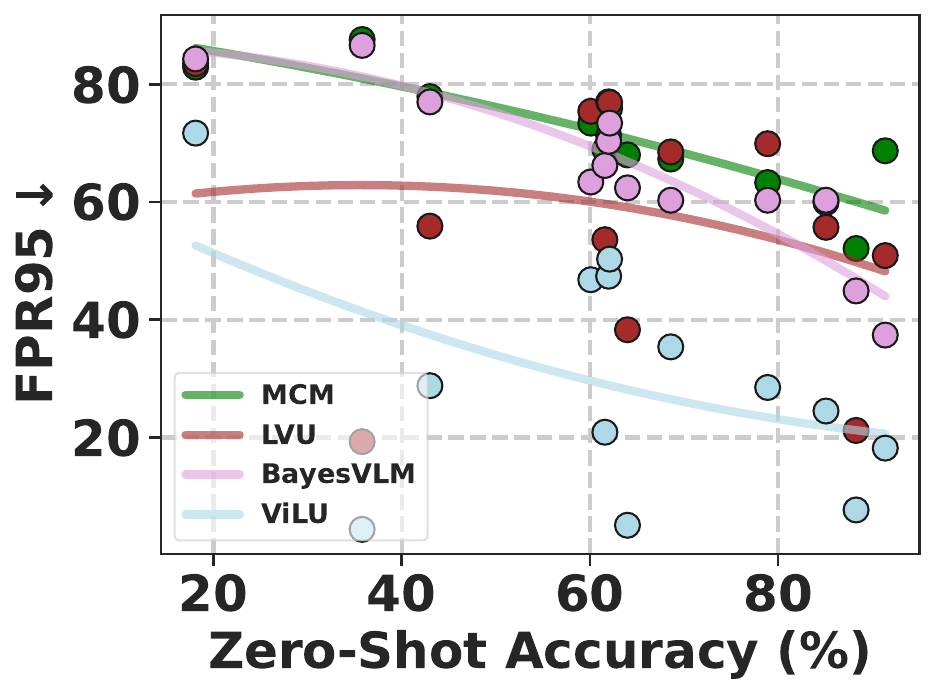} \\
        \end{tabular}}
        \caption{\textbf{Reliability of misclassification detection.} Our method, \ours, enhances misclassification detection by providing more reliable uncertainty estimates, particularly when zero-shot accuracy is low.}
        \label{fig:misclassification_zero_shot}
    \end{center}
\end{figure}

\cref{fig:misclassification_zero_shot} illustrates the relationship between misclassification detection performance and the zero-shot accuracy of the vision-language model for each baseline. Each dot at a given x-coordinate represents the classification performance of different baselines on the same dataset. The results emphasize the superior reliability of the uncertainty estimates provided by our method, particularly in low zero-shot accuracy settings. Notably, the tendency curves indicate a strong correlation between model performance and uncertainty metrics for both MCM and BayesVLM. Specifically, as zero-shot accuracy decreases, these two methods exhibit the worst performance. This suggests that they are only reliable when the model’s zero-shot accuracy is high—an unpredictable scenario in real-world settings, where ground-truth labels are unavailable.\\
\vspace{1em}

\subsection{Extension to different VLMs.}  
\label{sec:supp-backbones}
\cref{tab:backbone} presents the performance of \ours when applied to different zero-shot vision-language backbones, including CLIP~\cite{clip} and SigLIP~\cite{siglip}, with both ViT-B and ViT-L variants. Across all settings, \ours consistently outperforms MCM by a large margin in both AUC and FPR95, demonstrating strong and reliable failure detection. On CIFAR-10, the improvements are particularly pronounced: for example, using CLIP ViT-L/14, \ours achieves an AUC of 99.0 compared to 93.6 for MCM, and reduces FPR95 from 31.5 to just 4.1. On ImageNet-1k, the gains remain substantial, with up to 30-point reductions in FPR95. Unlike LVU-based methods~\cite{confidnet,Learning_loss_AL,pretrained_visual_uncertainties}, which require access to the model’s pre-training loss, \ours is trained solely from classification correctness, making it applicable to a broad range of frozen or proprietary VLMs. Overall, the consistent results across architectures confirm that \ours generalizes effectively with minimal assumptions.

\begin{table}[!h]
    \centering
    \small
    \setlength{\tabcolsep}{6pt}
    \resizebox{\linewidth}{!}{
    \begin{tabular}{lll cc  cc }
    \toprule
         \multicolumn{2}{c}{\multirow{2}{*}{Backbone}} & \multicolumn{1}{c}{\multirow{2}{*}{Method}} & \multicolumn{2}{c}{CIFAR-10} & \multicolumn{2}{c}{ImageNet-1k}  \\
         \cmidrule(lr){4-5} \cmidrule(lr){6-7} 
         & & & {\scriptsize AUC$\uparrow$} & {\scriptsize FPR95$\downarrow$} & {\scriptsize AUC$\uparrow$} &  {\scriptsize FPR95$\downarrow$} \\
         \midrule
         \multicolumn{1}{c}{\multirow{4}{*}{\rotatebox{90}{\textbf{\textbf{CLIP} \cite{clip}}}}} & \multirow{2}{*}{ViT-B/16} &
             MCM \cite{MCM}& 90.9  & 47.3  & 81.0  & 73.0   \\
         & & \ourcell \ours & \ourcell \textbf{98.4} & \ourcell \textbf{8.0} & \ourcell \textbf{90.3} & \ourcell \textbf{44.2}\\
         \cmidrule{2-7}
         & \multirow{2}{*}{ViT-L/14} & MCM \cite{MCM} & 93.6  & 31.5   & 82.9  & 68.9   \\
         & & \ourcell \ours & \ourcell \textbf{99.0} & \ourcell \textbf{4.1} & \ourcell \textbf{91.2} & \ourcell \textbf{39.2} \\
         \midrule
         \multicolumn{1}{c}{\multirow{4}{*}{\rotatebox{90}{\textbf{\textbf{SigLIP} \cite{siglip}}}}} & \multirow{2}{*}{ViT-B/16} &
             MCM \cite{MCM} & 92.8 & 46.1 & 84.2 & 65.8\\
         & & \ourcell \ours & \ourcell \textbf{97.6}& \ourcell \textbf{13.3}& \ourcell \textbf{90.7}& \ourcell \textbf{44.4}\\
         \cmidrule{2-7}
         & \multirow{2}{*}{ViT-L/16}  & MCM \cite{MCM} & 95.6 & 29.1&86.8 & 64.1\\
         & & \ourcell \ours & \ourcell \textbf{98.4}& \ourcell \textbf{7.8}& \ourcell \textbf{91.3}& \ourcell \textbf{41.4}\\
    \bottomrule
    \end{tabular}
    }
    \caption{\textbf{Generalization across backbones.} \ours shows consistent performance gains on several VLMs compared to MCM.}
    \label{tab:backbone}
\end{table}

\subsection{Domain generalization on ImageNet variants}
\vspace{1em}
To evaluate ViLU’s robustness under distribution shift, we consider a domain generalization setup in which ViLU is trained on the original ImageNet dataset and evaluated on two domain-shifted variants: ImageNet-V2 (IN-V2) and ImageNet-Sketch (IN-S). We first assess ViLU’s uncertainty estimates in a zero-shot transfer setting, where the model is applied directly to each variant without any adaptation. As shown in \cref{supp:table:imagenet_domain_gen}, ViLU achieves competitive performance, notably outperforming LVU on IN-S (FPR95 of 73.1 vs.\ 86.6) and remaining close to MCM (70.9). On IN-V2, ViLU performs even better, reaching an FPR95 of 54.8 compared to 71.7 for MCM and 77.3 for LVU. These results confirm that ViLU retains reliable uncertainty estimates even when evaluated on unseen domains.

We then explore a few-shot adaptation scenario, where ViLU is fine-tuned using only five labeled images per class from the target domain (IN-V2 or IN-S). On IN-S, this minimal supervision significantly reduces FPR95 from 73.1 to 54.4, outperforming both MCM (70.9) and LVU (72.3). On IN-V2, ViLU achieves similarly strong improvements, lowering FPR95 from 54.8 to 52.7, once again surpassing MCM (71.7) and LVU (68.7). These results highlight ViLU’s strong adaptability in low-data regimes and confirm that even minimal adaptation of the uncertainty head can lead to substantial gains in reliability under distribution shifts.\\

\begin{table}[!h]
\centering
\label{tab:imagenetv2_sketch}
\scriptsize
\begin{tabular}{lc|cc |cc}
\hline
 \textbf{Dataset} & \makecell{\textbf{MCM} \\[-0.75ex] \scriptsize{~}}  & \makecell{\textbf{LVU} \\[-0.75ex] \scriptsize{(\textit{zero-shot}})} & \makecell{\textbf{ViLU} \\[-0.75ex] \scriptsize{(\textit{zero-shot}})} & \makecell{\textbf{LVU} \\[-0.75ex] \scriptsize{(\textit{5-shot}})} & \makecell{\textbf{ViLU} \\[-0.75ex] \scriptsize{(\textit{5-shot}})}\\
\hline
\textbf{IN-V2} & 71.7 & 77.3 &  \ourcell{\textbf{54.8}} & 68.7 & \ourcell{\textbf{52.7}}\\
\textbf{IN-S}  & 70.9 & 86.6 &  \ourcell{73.1} & 72.3 & \ourcell{\textbf{54.4}}\\
\hline
\end{tabular}
\caption{FPR95$\downarrow$ on ViLU's domain generalization from ImageNet to ImageNet-V2 and ImageNet-Sketch.
}
\label{supp:table:imagenet_domain_gen}
\end{table}

\newpage
\subsection{Impact of concept coverage in pre-training}
\label{supp:sec:fpr_comparison}
In the main paper, we evaluated the zero-shot generalization ability of ViLU when pre-trained on CC12M and tested on 12 downstream datasets spanning various domains. In this section, we conduct a controlled experiment to assess whether better coverage of target concepts during pre-training improves zero-shot transfer. To this end, we construct a synthetic multi-dataset by combining the training sets of the 12 downstream datasets. Each image is paired with a pseudo-caption of the form “This is a photo of a ”, allowing us to train ViLU in the same image-caption setting as for CC12M. As shown in \cref{supp:tab:fpr_comparison}, this targeted pre-training leads to a substantial reduction in FPR95 across most datasets, with an average of 63.1 compared to 68.6 for the CC12M variant and 70.5 for MCM. These results confirm that more explicit coverage of the target classes during pre-training can significantly improve the quality of uncertainty estimates in zero-shot settings.

\begin{table}[!ht]
\centering
\footnotesize
\begin{tabular}{lc|c|c}
\hline
 & & \multicolumn{1}{c|}{\scriptsize{CC12M}} & \scriptsize{Multi-datasets}\\
\textbf{Dataset} & \textbf{MCM}  & \textbf{ViLU} & \textbf{ViLU}  \\
\hline
CIFAR-10        & 52.1 & \ourcell{54.2}               & \ourcell{\textbf{31.9}}   \\
CIFAR-100       & 67.3 & \ourcell{59.9}         & \ourcell{\textbf{50.3}}        \\
Caltech101      & 68.7 & \ourcell{\textbf{48.8}}      & \ourcell{70.8}           \\
Flowers102      & 68.0 & \ourcell{67.4}             & \ourcell{\textbf{45.9}}    \\
OxfordPets      & 59.9 & \ourcell{\textbf{58.1}}      & \ourcell{72.1}           \\
Food101         & 63.3 & \ourcell{67.4}              & \ourcell{\textbf{36.2}}   \\
FGVCAircraft    & 82.9 & \ourcell{82.3}              & \ourcell{\textbf{80.3}}    \\
EuroSAT         & 87.6 & \ourcell{\textbf{85.7}}       & \ourcell{91.3}          \\
DTD             & 77.9 & \ourcell{78.2}           & \ourcell{\textbf{75.2}}    \\
SUN397          & 75.9 & \ourcell{\textbf{72.7}}      & \ourcell{81.0}          \\
StanfordCars    & \textbf{73.4} & \ourcell{84.1}  & \ourcell{76.4}                 \\
UCF101          & 68.9 & \ourcell{63.8}      & \ourcell{\textbf{45.4}}       \\
\hline
\textbf{Average} & 70.5 & \ourcell{68.6}     & \ourcell{\textbf{63.1}}    \\
\hline
\end{tabular}
\caption{FPR95$\downarrow$ across datasets. Zero-shot performance when pre-trained on a Multi-datasets \vs CC12M.}
\label{supp:tab:fpr_comparison}
\end{table}

\subsection{Qualitative results}

\begin{figure}
    \centering
    \includegraphics[width=0.9\linewidth]{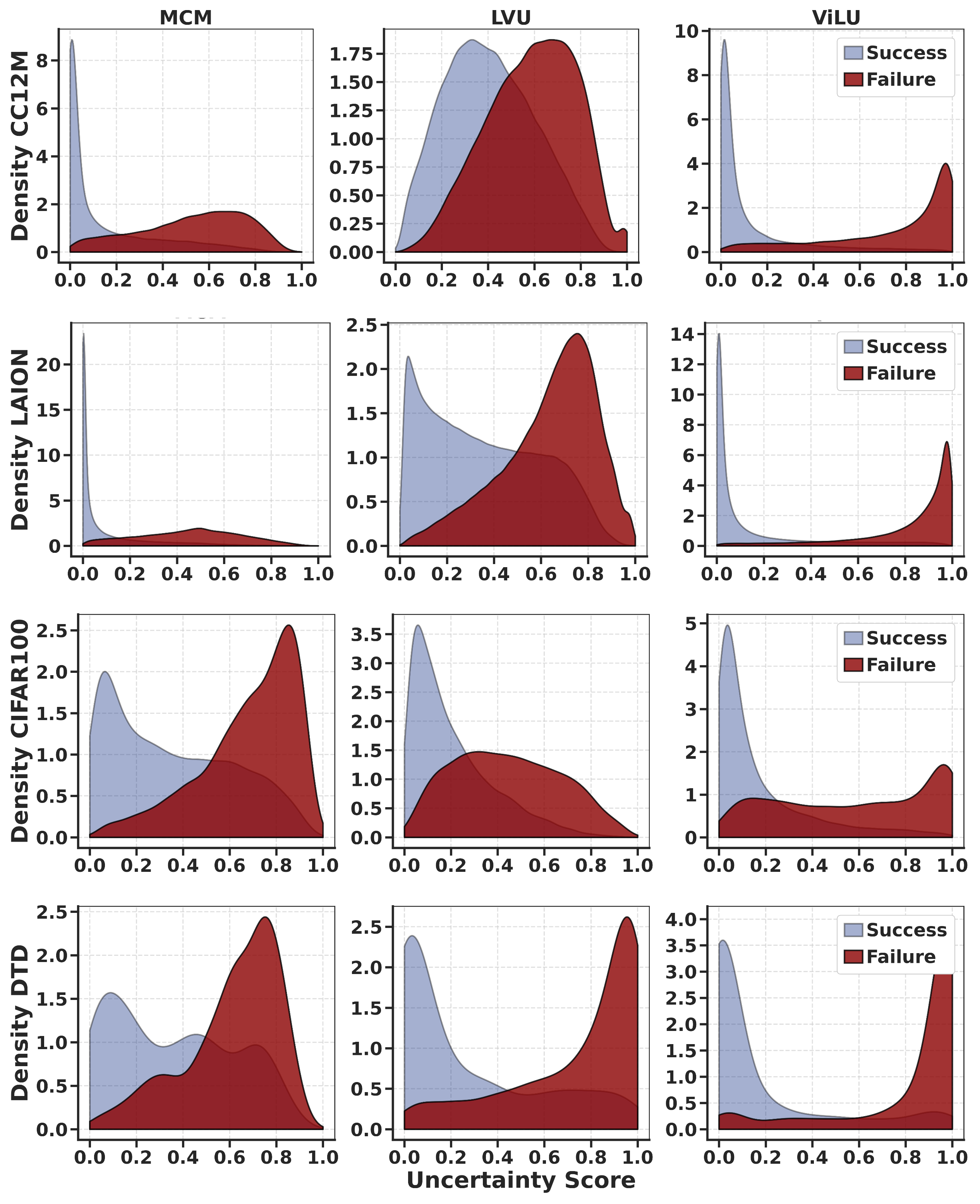}
    \caption{\textbf{Uncertainty score distribution.} Prediction for correctly and incorrectly classified samples on CC12M, LAION400M, CIFAR100 and DTD.}
    \label{fig:failure_curve_supp}
\end{figure}

\begin{figure}
    \centering
    \includegraphics[width=0.9\linewidth]{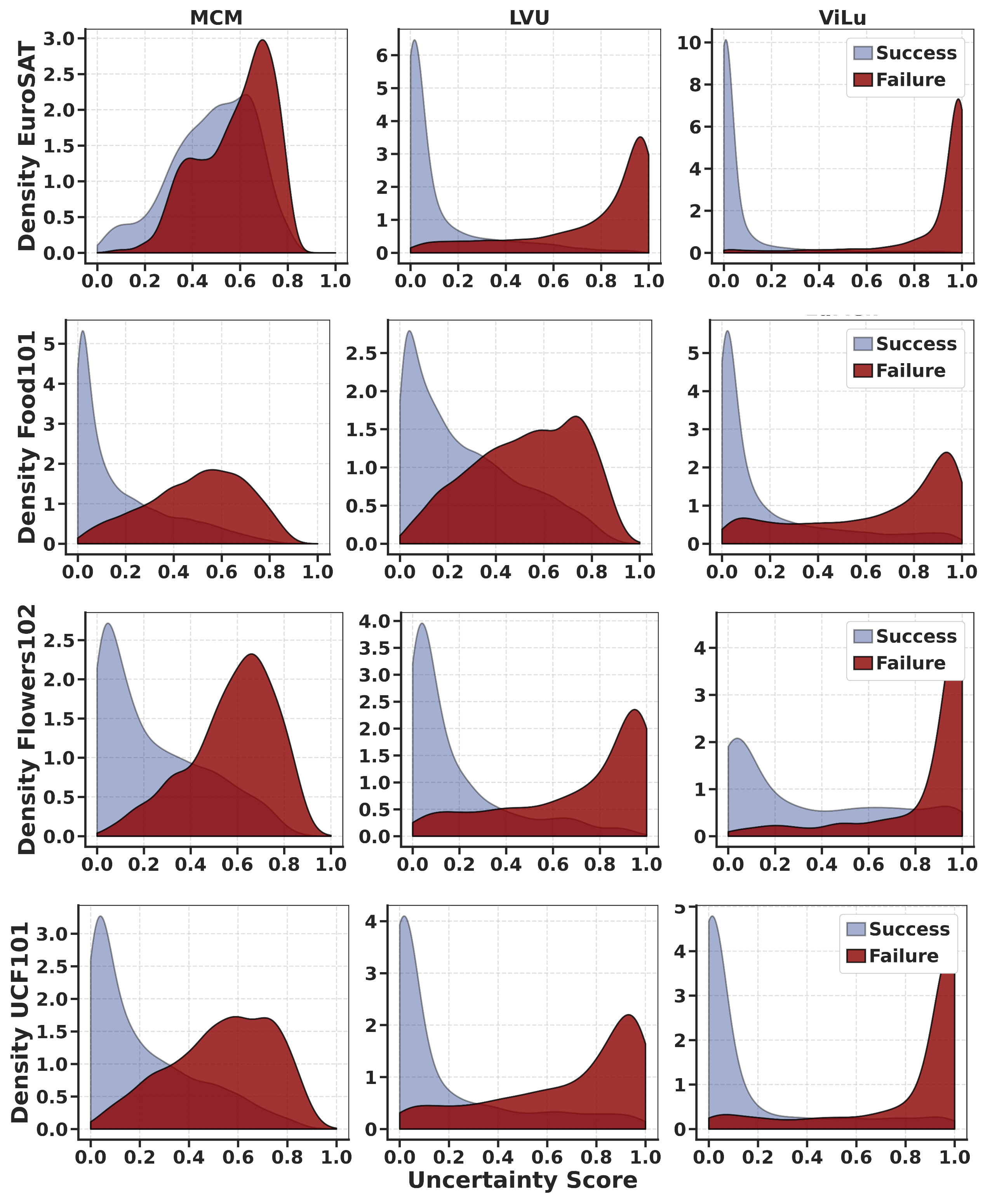}
    \caption{\textbf{Uncertainty score distribution.} Prediction for correctly and incorrectly classified samples on EuroSAT, Food101, Flowers102 and UCF101.}
    \label{fig:failure_curve1_supp}
\end{figure}

We provide additional visualizations on eight datasets in \cref{fig:failure_curve_supp} and \cref{fig:failure_curve1_supp}, illustrating the distribution of uncertainty scores for correctly and incorrectly classified validation samples. Our results demonstrate the consistency of \ours in assigning high uncertainty scores to misclassified samples (red) and low uncertainty scores to correctly classified ones (blue).

Unlike visual uncertainty models such as ConfidNet \cite{confidnet}, which rely solely on image features, our multimodal architecture leverages both visual and textual information to provide more reliable uncertainty estimates. Learning a cross-attention mechanism between image and text allows \ours to better capture ambiguities in class definitions, leading to improved uncertainty calibration across diverse datasets.


\clearpage
\end{document}